\pdfoutput=1

\documentclass[11pt]{article}

\usepackage[final]{acl}

\usepackage{times}
\usepackage{latexsym}

\usepackage[T1]{fontenc}

\usepackage[utf8]{inputenc}

\usepackage{microtype}

\usepackage{inconsolata}

\usepackage{graphicx}

\usepackage{amsmath}
\usepackage{amssymb}
\usepackage{url}
\usepackage{booktabs} 
\usepackage{graphicx}
\usepackage{epstopdf}
\usepackage{subfigure}
\usepackage{verbatim}
\usepackage{xcolor}         
\usepackage{bm}
\usepackage{array}
\usepackage{multirow}
\usepackage{subfigure}
\usepackage{makecell}
\usepackage{xcolor}
\usepackage{xspace}
\usepackage{framed}
\usepackage{changepage}
\usepackage{cleveref}
\usepackage{hyperref}
\usepackage{enumitem}
\usepackage{hyperref}
\usepackage{siunitx}
\usepackage{tablefootnote}
\usepackage{soul}
\usepackage{xcolor}
\usepackage{colortbl}

\definecolor{lightpurple}{RGB}{230, 190, 255}
\definecolor{lightblue}{rgb}{0.8,0.85,1}
\definecolor{lightgreen}{rgb}{0.8,1,0.8}
\definecolor{lightyellow}{rgb}{1,1,0.8}
\definecolor{lightred}{rgb}{1,0.8,0.8}

\sethlcolor{lightpurple}

\usepackage[most]{tcolorbox}

\usepackage[font={small}]{caption}

\newcommand{\zhiyu}[1]{\textcolor{red}{[zhiyu: #1]}}

\newcommand{\patient}{\textsc{Patient-$\Psi$}\xspace}
\newcommand{\trainer}{\textsc{Patient-$\Psi$-Trainer}\xspace}
\newcommand{\data}{\textsc{Patient-$\Psi$-CM}\xspace}

\definecolor{formalshade}{rgb}{0.8, 0.9, 0.8}
\newenvironment{formal}{%
  \MakeFramed{\advance\hsize-\width\FrameRestore}%
  \noindent\hspace{-4.55pt}
  \begin{adjustwidth}{}{7pt}%
  \vspace{0.5pt}\vspace{0.5pt}%
}
{%
  \vspace{0.5pt}\end{adjustwidth}\endMakeFramed%
}

\newenvironment{contributionenum}
{%
    \begin{itemize}[itemsep=0.5pt, parsep=0pt, topsep=0.5pt, partopsep=0pt]
}
{%
    \end{itemize}
}

\newlist{researchqs}{enumerate}{1}
\setlist[researchqs,1]{
    label=RQ \arabic*,
    leftmargin=*,
    itemsep=0.2em,
    parsep=0em,
    topsep=0.2em,
    partopsep=0em
}

\title{\textsc{Patient-$\Psi$}: Using Large Language Models to Simulate Patients for Training Mental Health Professionals}

\author{\textbf{Ruiyi Wang}\textsuperscript{*}\textsuperscript{1}, 
\textbf{Stephanie Milani}\textsuperscript{*}\textsuperscript{1}, 
\textbf{Jamie C. Chiu}\textsuperscript{2},
\textbf{Jiayin Zhi}\textsuperscript{1}, \\
\textbf{Shaun M. Eack}\textsuperscript{3}, 
\textbf{Travis Labrum}\textsuperscript{3}, 
\textbf{Samuel M. Murphy}\textsuperscript{3},
\textbf{Nev Jones}\textsuperscript{3},
\textbf{Kate Hardy}\textsuperscript{4}, \\
\textbf{Hong Shen}\textsuperscript{1},
\textbf{Fei Fang}\textsuperscript{1},
\textbf{Zhiyu Zoey Chen}\textsuperscript{1}
\\
  \textsuperscript{1}School of Computer Science, Carnegie Mellon University, \\
  \textsuperscript{2}Department of Psychology, Princeton University, \\
  \textsuperscript{3}School of Social Work, University of Pittsburgh, \\
  \textsuperscript{4}Department of Psychiatry and Behavioral Sciences, Stanford University \\
  {\tt \{ruiyiwan,smilani\}@andrew.cmu.edu}, \tt zhiyu.chen2@utdallas.edu}

\begin{document}
\maketitle

{\let\thefootnote\relax\footnotetext{\hspace{-1.5mm}*\hspace{0.2mm}Major contributors. See \S\ref{app:contirbutions} for individual contributions.}}

\begin{abstract}
Mental illness remains one of the most critical public health issues.  
Despite its importance, many mental health professionals highlight a disconnect between their training and actual real-world patient practice. 
To help bridge this gap, we propose \patient, a novel patient simulation framework for cognitive behavior therapy (CBT) training. To build \patient, we construct diverse patient cognitive models based on CBT principles and use large language models (LLMs) programmed with these cognitive models to act as a simulated therapy patient. We propose an interactive training scheme, \trainer, for mental health trainees to practice a key skill in CBT -- formulating the cognitive model of the patient -- through role-playing a therapy session with \patient.
To evaluate \patient, we conducted a comprehensive user study of 13 mental health trainees and 20 experts. 
The results demonstrate that practice using \trainer enhances the perceived skill acquisition and confidence of the trainees beyond existing forms of training such as textbooks, videos, and role-play with non-patients. Based on the experts' perceptions, \patient is perceived to be closer to real patient interactions than GPT-4, and \trainer holds strong promise to improve trainee competencies. 
Our code and data are released\footnote{\url{https://github.com/ruiyiw/patient-psi}}.

\end{abstract}

\section{Introduction}
\begin{figure}[ht]
\centering
\includegraphics[width=0.48\textwidth]{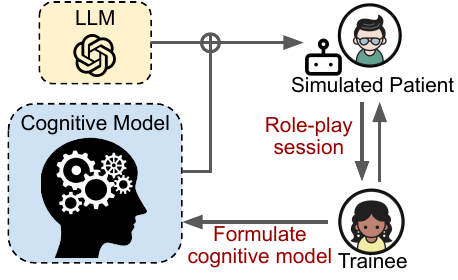}
\caption{Illustration of our patient simulation idea.} 
\label{fig:cog_model}
\end{figure}
One in eight people globally are living with mental health conditions (World Health Organization, 2023)\footnote{\url{https://www.who.int/campaigns/world-mental-health-day/2023}}. However, there is a significant gap between the available mental health support and patient needs, with over half (54.7\%) of adults with a mental illness receiving no treatment in the US\footnote{\url{https://mhanational.org/issues/2023/mental-health-america-access-care-data}}.
Training mental health professionals requires extensive effort, yet many professionals highlight a disconnect between their training and the complexities of real patient interactions. 
To understand these training challenges, we conducted a formative study involving semi-structured interviews with twelve mental health experts and trainees.
This diverse group comprised of clinical psychologists, licensed social workers, and current master's students in social work. 
The experts provided insights into the difficulties faced when transitioning from formal CBT training to real-world practice (details in~\Cref{appx:formative_study}).
All experts noted that their training did not adequately prepare them for the unpredictable and multifaceted nature of real patient interactions.
Despite wanting more interactive experiences, they found role-playing exercises with peers, a common training method, to be unrealistic, as these exercises often do not reflect actual therapy sessions. 

There has been growing interest in developing LLM-based methods for psychology~\citep{demszky2023,chen2023empowering}. In  \citep{bubeck2023sparks,kosinski2023theory}, ChatGPT and GPT-4 are able to solve some basic theory of mind tasks that generally require the ability to understand and attribute mental states to oneself and others. 
Inspired by such promise, we propose to use LLMs to simulate patients to train mental health professionals, with the goal of bridging the gap between their existing training methods and the complexities of real patient interactions. However, two major challenges must be addressed to realize this idea:

\noindent \textbf{Fidelity.} \textit{How can we build simulated patients that closely resemble the communicative behaviors of real patients with mental health disorders?}

\noindent \textbf{Effectiveness.} \textit{How can we design an effective training scheme that allows trainees to benefit from interacting with these simulated patients?}

In this work, we claim that integrating a patient's \textit{cognitive model} with an LLM can achieve high fidelity in simulating real patients with mental health disorders corresponding to that cognitive model. We implement this idea using the cognitive modeling framework in CBT~\citep{beck2020cognitive}, a popular paradigm in psychotherapy.
We propose \patient, a novel simulated patient agent that integrates cognitive modeling with LLMs. We collaborate with clinical psychologists to curate a dataset, \data, which comprises 106 high-quality and diverse patient cognitive models. 
These cognitive models cover unhealthy cognitive structures embedded in multiple contexts, such as family issues, relationship problems, workplace challenges, and more.
We then use these cognitive models to program an LLM to act as the \patient agent. To better resemble the complex dynamics of real patient communications within a therapy session, we also integrate six conversational styles into \patient. These conversational styles were identified from our formative study with mental health domain experts. 

In CBT, formulating a patient's cognitive model is a crucial skill that therapists need to learn~\cite{beck2020cognitive}. Our design of \patient naturally incorporates a feedback mechanism for trainees to practice this skill without extensive need for supervisor intervention, which is a desired benefit of AI-based training. We propose \trainer, an interactive training framework for mental health trainees to practice CBT cognitive model formulation using \patient. Specifically, trainees converse with the simulated patient, \patient, to formulate its cognitive model. Afterward, the system displays the original cognitive model that was used to program the simulated patient as a \textit{reference}, allowing trainees to compare their results as feedback. Within this training framework, the effectiveness of the feedback theoretically depends on how accurately \patient simulates a real patient with the corresponding cognitive model. Figure~\ref{fig:cog_model} illustrates the overall idea of our framework.

To evaluate the fidelity of \patient and the effectiveness of \trainer, we conducted a thorough user study with \textbf{20 mental health experts} and \textbf{13 trainees}. Evaluation results from the experts indicate that: (1) \patient closely resembles real patients in terms of maladaptive cognitions, conversational styles, and emotional states; outperforming GPT-4. (2) Practicing with \trainer is perceived to be highly beneficial for improving CBT formulation skills and better-preparing trainees for interactions with real patients. Experts also highlighted several advantages of \trainer, including customized options to choose conversation styles and the diverse patient cognitive models. Evaluation results from the trainees indicate that practicing with \trainer is perceived to improve skill and confidence, compared to current training methods. Overall, experts and trainees prefer using \trainer over a strong GPT-4 baseline. We also demonstrate that automatic evaluations with LLMs fail to assess the simulated patient fidelity, indicating the challenge of our task. 
Our contributions are summarized as follows:
\begin{contributionenum}
    \item We propose \patient, a novel simulated therapy patient, built using cognitive models grounded in psychology principles and LLMs.
    \item We propose \trainer, an interactive training framework for trainees to practice CBT formulation skills on \patient . 
    \item We create and release a dataset, \data, with high-quality CBT-based cognitive models curated by clinical psychologists.
    \item Our user study with both mental health trainees and experts demonstrates that \patient exhibits high fidelity to real patients, and practicing with \trainer significantly improves perceived skills and confidence in CBT formulation.
\end{contributionenum}

\section{Methodology}
\begin{figure*}[t]
\centering
\includegraphics[width=1.0\textwidth]{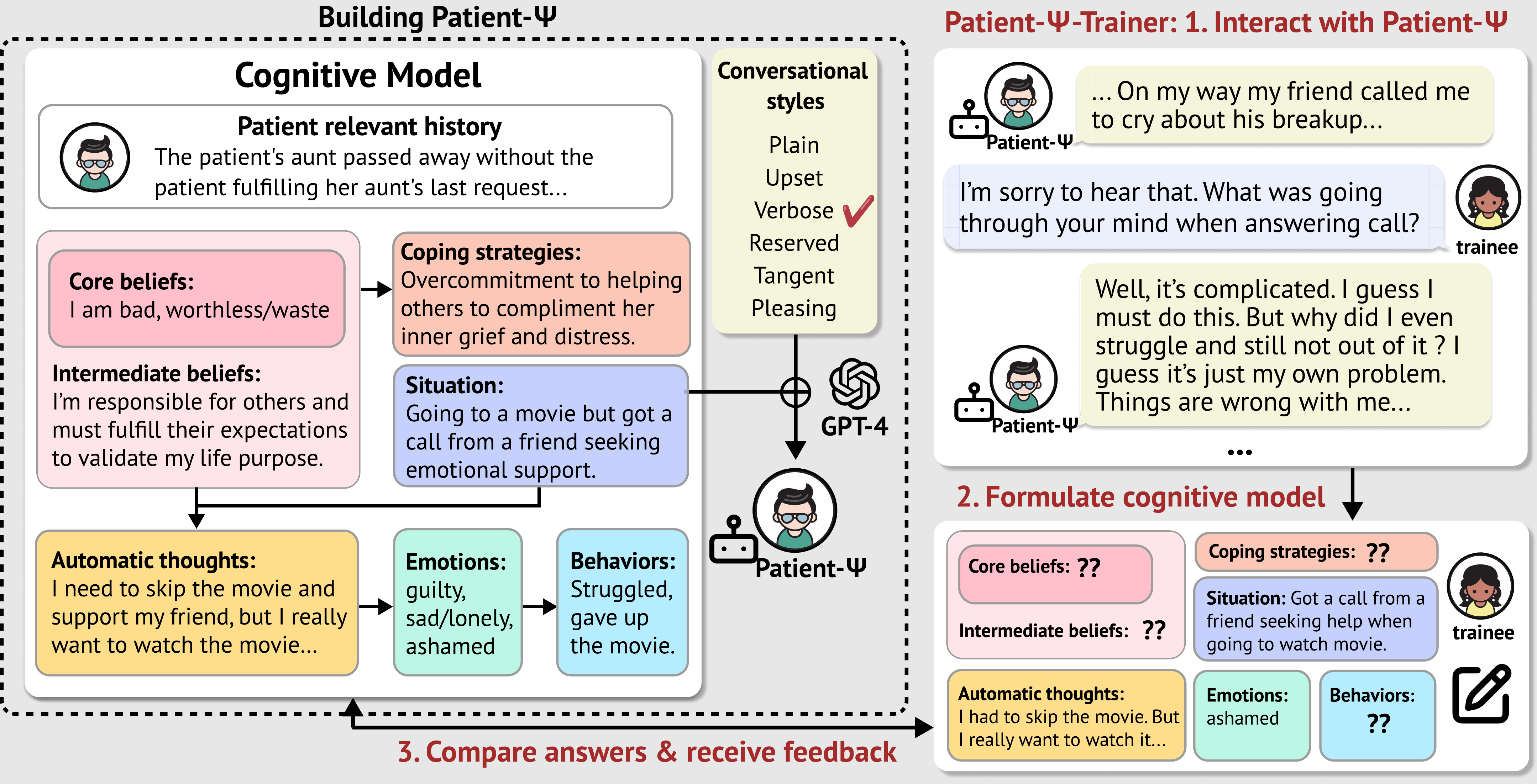}
\caption{The overall framework of \patient and \trainer. We integrate the expert-created cognitive model with GPT-4 to build \patient. In \trainer, the trainee role-plays a therapy session with \patient to formulate its cognitive model. The trainee can compare their formulation with the cognitive model used to build \patient to get feedback.} 
\label{fig:overview}
\end{figure*}
In this section, we first describe the construction of \patient in \S\ref{method_patient}. We detail the integration of \emph{cognitive models} with LLMs and the incorporation of \emph{conversational styles} to accurately mimic real patient interactions. Next, we explain the training framework, \trainer, in \S\ref{method_trainer}, which utilizes \patient to create an interactive learning environment for practicing CBT formulation skills. \Cref{fig:overview} provides an overview of our method.

\subsection{\patient}
\label{method_patient}

\paragraph{Using Cognitive Models to Simulate Patients.}
\emph{Cognitive models} in mental health provide a structured framework for understanding how an individual's thoughts and beliefs are interconnected and influence emotions and behaviors. In established therapy paradigms like CBT~\cite{beck2020cognitive}, formulating a patient's cognitive model is central for a therapist to understand and address the maladaptive cognitions maintaining distress and symptoms~\cite{hollon2013cognitive,hofmann2012efficacy}.

The \textit{Cognitive Conceptualization Diagram} (CCD)~\cite{beck2020cognitive} is a commonly used representation of a patient's cognitive model in CBT.
The left side of \Cref{fig:overview} depicts an example CCD-based cognitive model, illustrating eight key components.
{\large \textcircled{\small 1}} \textit{Relevant history} contains significant past events that contribute to an individual's mental state.
{\large \textcircled{\small 2}} \textit{Core Beliefs} are deeply ingrained perceptions about oneself, others, and the world.
{\large \textcircled{\small 3}} \textit{Intermediate beliefs} are the underlying rules, attitudes, and assumptions derived from core beliefs and shape an individual's thought patterns. 
{\large \textcircled{\small 4}} \textit{Coping strategies} are techniques used to manage negative emotions.
An external event or context ({\large \textcircled{\small 5}} \textit{a situation}) may trigger quick, evaluative thoughts without deliberation ({\large \textcircled{\small 6}} \textit{automatic thoughts}) stemming from the beliefs, leading to responses in terms of {\large \textcircled{\small 7}} \textit{emotions} and {\large \textcircled{\small 8}} \textit{behaviors}. 
A CCD-based cognitive model links the components together, creating a framework for identifying and understanding patients' underlying cognitive processes. 
For all the components, we adopt the definitions and formulations put forth by \cite{beck2020cognitive}. These include: three major core beliefs ({\large \textcircled{\small 2}})—helpless, unlovable, and worthless—each with several fine-grained core beliefs, for a total of 19 core belief categories; 9 emotion ({\large \textcircled{\small 7}}) categories; the rest of the components are formulated as free-text entries.
See \Cref{tab:dataset_statistics} and \Cref{appx:patient_details_ccd} for the categories.
In this work, we integrate CCD-based cognitive models into an LLM to simulate patients whose communication reflects the underlying cognitive processes.

\paragraph{The \data Cognitive Model Dataset.}
To the best of our knowledge, no existing work offers a dataset of realistic cognitive models due to two challenges: 1) the data privacy constraints involved in acquiring real patient cognitive models and 2) the high-level expertise required to perform manual creations. 
In this work, we propose the first dataset of CCD-based cognitive models grounded in CBT principles, \data, created by clinical psychologists.
We first prompt GPT-4 Turbo~\citep{openai2023gpt4} to create summaries from therapy session transcripts. 
These transcripts were obtained from the Alexander Street database\footnote{\url{https://alexanderstreet.com/}, accessed through our institution's subscription.} under the subject ``Counseling and Therapy'' and the keyword ``Cognitive Behavioral Therapy''. 
Two clinical psychologists then manually create cognitive models by drawing inspiration from the transcript summaries, incorporating their professional expertise, and applying their creativity (within clinical constraints). 
This process involves developing new cases inspired by the summaries and composing the corresponding cognitive models.
We emphasize \textit{diversity} and \textit{realism} to the psychologists when creating the models. We end up with a dataset containing 106 cognitive models (an example is shown in \Cref{fig:overview}, left). Each cognitive model is associated with one activating situation. See \Cref{appx:patient_details_dataset} for details of dataset creation and more example cognitive models from \data.


\begin{table}[t]
\centering
\resizebox{0.35\textwidth}{!}{
\begin{tabular}{>{\ttfamily}l p{4.5cm}}
\toprule
\textnormal{Style} & Description \\
\midrule
\rowcolor{lightblue} plain & Direct, straightforward. \\
\rowcolor{lightyellow} upset & Frustration, resistance. \\
\rowcolor{lightyellow} verbose & Overly expressive.  \\
\rowcolor{lightyellow} reserved & Minimal, restrained. \\
\rowcolor{lightyellow} tangent & Deviates from the main topic. \\
\rowcolor{lightyellow} pleasing & Agreeable to a fault. \\
\bottomrule
\end{tabular}
}
\caption{Different conversational styles that \patient can take on, with descriptions. 
More detailed examples in \Cref{appx:patient_details_conversational}. 
\colorbox{lightyellow}{Yellow styles} are harder; \colorbox{lightblue}{blue style} is easier. 
}
\label{tab:relational_styles}
\end{table}

\begin{table}[t]
\small
\begin{tabular}{lc|llc}
\toprule
\textbf{Situations}    & \textbf{\#} &  & \textbf{Emotions} & \textbf{\#} \\
\midrule
family dynamics        & 25          &  & anxious           & 60          \\
workplace pressure     & 20          &  & sad               & 50          \\
relationship dynamics  & 19          &  & angry             & 22          \\
social interactions    & 18          &  & hurt              & 21          \\
personal growth issues & 8           &  & disappointed      & 19          \\
financial concerns     & 8           &  & ashamed           & 17          \\
daily life stressors   & 8           &  & guilty            & 13          \\ 
\multicolumn{2}{l}{\rule{0.5\linewidth}{0.4pt}} & & suspicious & 2\\
\textbf{Core beliefs}  & \textbf{\#} &  & jealous & 1       \\
helpless               & 94          &  &            &           \\
unlovable              & 71          &  &                   &             \\
worthless              & 15          &  &  \multicolumn{2}{l}{\textbf{106 cognitive models}}             \\
\bottomrule
\end{tabular}
\caption{\data statistics. Details in \Cref{appx:patient_details_ccd}.}
\label{tab:dataset_statistics}
\end{table}

\paragraph{Conversational Styles Integration.}
In the formative study (\Cref{appx:formative_study}), domain experts emphasized that real patients exhibit different \textit{conversational styles} during therapy. Based on these discussions, we identify six styles for \patient, 
detailed in \Cref{tab:relational_styles}.
To create a natural curriculum, the styles are two levels of difficulty.
The easiest style, \texttt{plain}, features direct and straightforward communication.
The more difficult styles require trainees to exert more effort to elicit relevant information.
To incorporate these styles with \patient, two clinical psychologists annotate each cognitive model with a list of valid conversational styles and develop instructions for \patient to simulate a patient for each style. Detailed descriptions and examples of the conversational styles are provided in \Cref{appx:patient_details_conversational}.

\paragraph{Patient Agent Simulation.}
We prompt GPT-4 to build \patient which consists of a patient's cognitive model, the conversational style prompt, and a list of instruction prompts. 
\Cref{appx:patient_details_prompts} contains the full prompts.
The model is continually prompted to engage in a CBT-based therapy session, role-playing a patient with the corresponding cognitive model and conversational styles.

\subsection{\trainer}
\label{method_trainer}
With the development of \patient, we introduce \trainer, an interactive training framework designed for mental health professionals to practice cognitive model formulation for CBT. 
\trainer offers a structured, three-step training process: 1) engaging with \patient in a simulated CBT session, 2) formulating \patient's cognitive model through interaction, and 3) reviewing the original cognitive model used to create \patient for feedback. 
The right-hand side of \Cref{fig:overview} illustrates this process.

\paragraph{Training Process.}
Trainees begin by choosing one of the six conversational styles from \trainer's web application interface (screenshots in \Cref{app:inter}). 
Then, a patient is generated using the chosen style and a randomly-selected cognitive model from \data compatible with that style. The interface displays the patient's relevant history in preparation for the session. During this session, the trainee engages with \patient, applying their therapeutic skills with the goal of formulating the CCD-based cognitive model used to program \patient. 
This involves eliciting and summarizing 
all cognitive elements underlying the conversation with \patient.

\paragraph{Real-Time Feedback.}
Upon concluding the interactive session, \trainer allows the trainee to compare their formulated cognitive model with the original cognitive model used to program \patient. This side-by-side comparison highlights discrepancies, providing detailed feedback. Trainees can continue to chat with \patient to refine their formulations. This natural feedback loop, stemming from our design of using the cognitive model to program the patient, offers the advantage of minimal human supervision efforts, enabling trainees to practice independently.

\section{Experiment Setup}
We now present the experimental setup for evaluating \patient and \trainer.
We aim to answer the following research questions:
\begin{researchqs}
    \item \textbf{Fidelity:} Does \patient improve the fidelity of patient simulations compared to baselines?
    \item \textbf{Accuracy:} How closely does \patient imitate the underlying cognitive model? 
    \item \textbf{Effectiveness:} Do experts and trainees perceive \trainer as an effective tool for CBT training?
    \item \textbf{AutoEval:} Can we leverage existing methods, such as LLMs, to automatically evaluate the patient simulations?
\end{researchqs}
In \S\ref{sec:user_study_results}, we answer the first three RQs through our user study with both trainees and experts.
Then, in \S\ref{sec:auto_eval_results}, we show that current automatic evaluations cannot capture the nuances necessary for conducting highly technical, domain-specific assessments.
This finding not only shows the importance of user study evaluations but also motivates future work on performant automatic evaluators.

\paragraph{Evaluation Dimensions.}

We design a set of fine-grained dimensions to assess each RQ, using insights from the formative study and existing literature \citep{beck2020cognitive, Bouter2012, Issenberg2005, Silverman2013, Ekman1992AnAF}. 
To ensure that the simulated patients' responses reflect those of real patients, we measure the \textbf{fidelity} of the \textit{emotional states}, \textit{conversational styles}, and \textit{maladaptive cognitions} of \patient to real patients.
To assess the \textbf{accuracy} of \patient in emulating the underlying expert-validated cognitive model, we evaluate each component's accuracy. 
To assess the \textbf{effectiveness} of \trainer, 
we measure the perceived improvements of CBT formulation skills: \textit{identifying maladaptive thinking patterns} and \textit{identifying beliefs}. 
We also measure the perceived \textit{confidence improvement} of the trainees. 
Finally, we assess \textit{usability} to ensure the tool's ease of use for users. Due to space constraints, the usability results are in \Cref{app:subsec_usability}. 

For pairwise comparisons, the options are: ``A is much better than B," "A is somewhat better than B," "about the same," "B is somewhat better than A," and "B is much better than A." 
We map the results to a scale from -2 to 2, where $\pm 2$ indicates a strong preference. 
Individual measures use a 5-point Likert scale from 1 to 5,  
where $5$ means "strongly agree" or "extremely accurate," and $1$ means "strongly disagree" or "not accurate at all." Specific values for each dimension are in \S\ref{sec:user_study_results}.

\paragraph{Baselines.} 
We leverage vanilla GPT-4 with a general description of patients with depression or anxiety as the input, rather than the cognitive models (see \Cref{sec:user_study_design_details}).
Thus, we cannot show the reference cognitive model as feedback and do not include the conversational styles.
We also compare with existing training techniques, which includes peer role-plays or textbook examples.


\paragraph{User Study Details.}

Assessing simulated therapeutic dialogue is a cognitively difficult process that requires professional training and experience, making typical crowdsourcing data collection approaches difficult.
To ensure high-quality evaluations from those with significant real patient experience (experts) and from the population who would use \trainer in practice (trainees), we recruit 20 current mental health practitioners and 13 social work students, respectively.\footnote{We recruited participants through the professional networks of our co-authors in mental health (clinical psychologists and professors in clinical psychology and social work), as well as snowball sampling.}
\S\ref{ethics} details the IRB approval and recruitment.
Each participant practices with \trainer and the baseline in a randomized order, completing two simulated patient sessions for each. To ensure comprehensive evaluation across diverse cognitive models, we assign each participant simulated patients with distinct underlying cognitive models, covering a total of 66 cognitive models from \data. For expert evaluations, we distribute two specific conversational styles to each participant to achieve an overall balanced distributions of all styles. Trainees can select two styles based on their expertise level and confidence.
More protocol details are in \Cref{sec:user_study_design_details}.

\paragraph{Automatic Evaluation Details}
To leverage LLMs' capabilities as a judge for evaluating open-ended tasks, we use state-of-art LLMs, including GPT-4~\citep{openai2023gpt4} and Llama 3 70B~\citep{llama3modelcard}, as evaluators. We set the temperature to 1.0 for both LLMs. For assessing fidelity, we prompt LLM evaluators with the same instructions used in the questionnaire and the conversations between experts and \patient, as collected during the user study. The Likert scale ratings are mapped to numerical values ranging from 1 to 5 to present the results in a direct, interpretable format. 
For evaluating accuracy, we apply two approaches based on the types of the CBT components being assessed: 
\begin{itemize}
    \item[1.] \textbf{Text-based fields} (including situation, coping strategies, intermediate beliefs, automatic thoughts, and behaviors): we generate a multiple-choice question by randomly sampling four additional components from \data, alongside the ground-truth component. GPT-4 is then prompted to select the component most closely reflected in \patient's conversations. We report the accuracy based on a 5-class classification task.
    \item[2.] \textbf{Categorization fields} (including core beliefs and emotions): These components are selected from predefined sets of core beliefs and emotions as described in CBT literature~\citep{beck2020cognitive}. GPT-4 is prompted to choose the options that are reflected in the patient's conversations, and we report the F1 score for these selections. 
\end{itemize}

\section{User Study Results}
\label{sec:user_study_results}
\begin{figure*}[t]
    \centering
    \includegraphics[width=\textwidth]{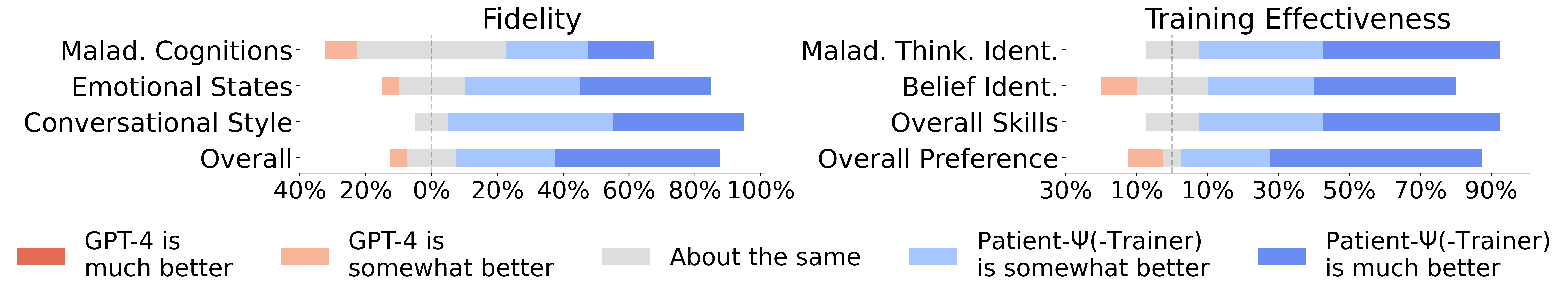}
    \vspace{-1.8em}
    \caption{{Fidelity of \patient and training effectiveness of \trainer compared to GPT-4 baseline along multiple dimensions. 
    X-axis: the \% of experts who voted for a specific option; y-axis: the assessment dimension.
    Malad. means maladaptive, Think. means thinking, and Ident. means identification.
    \patient more closely resembles real patients (fidelity, left) and is considered more effective for trainees (training effectiveness, right).}}
    \label{fig:fidelity_effectiveness}
\end{figure*}

\subsection{RQ 1: Fidelity to Real Patients}
\label{sec:user_study_fidelity}
To assess the fidelity of \patient to real patients, experts compare existing training techniques, the GPT-4 baseline, and \patient
(\Cref{tab:fidelity_experience}).
We ask experts for their overall impressions of these training methods, resulting in 20 data points for each comparison in this subsection.
Paired t-tests show that \patient significantly outperforms the other methods ($p < 10^{-4}$), indicating that \patient provides the most realistic patients, addressing RQ 1 positively.
This is promising for \patient: our formative study highlighted a gap in trainee preparation for real interactions, which \patient can effectively fill.

\begin{table}[t]
    \centering
    \resizebox{0.43\textwidth}{!}{
    \begin{tabular}{llc}
    \toprule 
       Comparison  &  Patient Fidelity $\mu$ \\ 
    \midrule 
        \textbf{\patient} vs. GPT-4 & \num{1.25}*** \\
        \textbf{\patient} vs. Traditional & \num{1.25}*** \\
        \textbf{GPT-4} vs. Traditional & \num{0.7}* \\ 
    \midrule 
    \multicolumn{2}{l}{* : $p < 0.05$, ** : $p < 0.01$, *** : $p < 10^{-4}$ } \\
    \bottomrule
    \end{tabular}}
    \vspace{-0.3em}
    \caption{\patient provides significantly more realistic simulated patients compared to the GPT-4 baseline and traditional methods. 
    Closer to 2/-2: the first/second method is better.
    }
    \label{tab:fidelity_experience}
\end{table}

\paragraph{\patient exhibits higher fidelity to real patients than the GPT-4 baseline.}
Each expert compares the fidelity dimensions (\textit{emotional states}, \textit{conversational styles},  \textit{maladaptive cognitions}) of \patient and the GPT-4 baseline to real patients.
\Cref{fig:fidelity_effectiveness} (left) depicts the distribution of expert comparisons; summary statistics in \Cref{tab:fidelity_dims_decomposed}, \Cref{appx:user_study}.
\patient{} is rated higher along all dimensions for fidelity: it better represents the maladaptive cognitions ($\mu$: $\num{0.55}$, 
$p<0.05$), 
the emotional states ($\mu$: $\num{1.1}$, 
$p<10^{-4}$), 
and the conversational styles ($\mu$: $\num{1.3}$, 
$p < 10^{-4}$)
of real patients. 
Experts expressed that \patient offered a more realistic challenge of extracting information from patients, unlike the baseline which was too forthcoming with responses. 
One expert noted that sessions with the baseline felt ``almost like doing therapy with a therapist,'' highlighting the challenge of simulating real patient behavior --- even with advanced LLMs likely pretrained on an extensive corpus of therapy knowledge.


\subsection{RQ 2: Accuracy to Cognitive Model} 
\label{sec:user_study_accuracy}
\begin{figure}[t]
    \centering
    \includegraphics[width=\columnwidth]{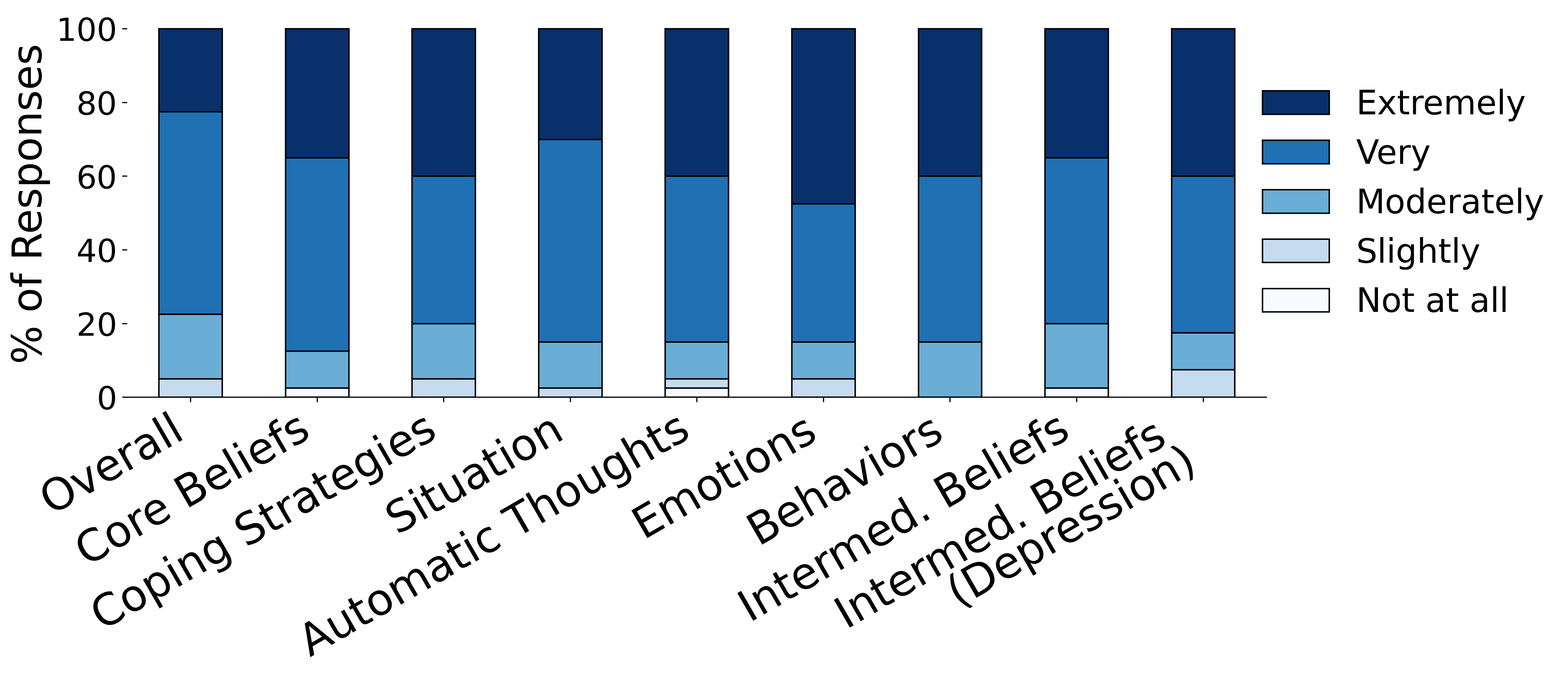}
    \vspace{-1.8em}
    \caption{Experts rate 97\% of the \patient patients as at least moderately accurate in reflecting the reference cognitive model. Intermed. means Intermediate. 
    }
    \label{fig:results_accuracy}
\end{figure}

To be practically useful, \patient must accurately reflect the reference cognitive model, as trainees rely on it for feedback on their completed formulations. 
Experts evaluate \patient's overall accuracy and its accuracy for each component of the cognitive model, resulting in $40$ data points per dimension. 
\Cref{table:mean_accuracies}, \Cref{appx:subsec_accuracy} presents the summary statistics; \Cref{fig:results_accuracy} illustrates the distribution. 
The results are promising: overall, \patient is rated on average as \textit{very accurate}. 
For each of the 8 components, \patient is rated on average as \textit{very} to \textit{extremely} accurate. 
Specifically, 80-88\% of the simulated patients achieve \textit{very} to \textit{extremely accurate} ratings for each of the 8 components, answering RQ 2. 
Crucially, since the reference cognitive model is accurately captured by \patient, trainees can rely on it to receive high-quality feedback on their responses.

\paragraph{Patient simulation may involve an accuracy-fidelity trade-off.}
Expert feedback from our study reveals insights into the challenges of accurately simulating patients in alignment with the underlying cognitive model. 
Specifically, there exists a tension between some of the evaluation metrics.
For example, one expert, noting the limitations of text-only interfaces, suggests increasing the use of explicit emotion words to improve the model's ability to accurately convey emotions. 
However, this approach potentially conflicts with real-world language patterns, as highlighted by another expert. 
This expert works with populations from prisons.
According to the expert, this population tends to not use any feeling words, so the expert believes that including emotion words is less realistic based on their experience with this subpopulation. 
As a result, including such words may improve accuracy but could do so at the cost of fidelity. 
This finding suggests that, in some cases, there exists a tension between evaluation metrics.

\subsection{RQ 3: Effectiveness for Training}
\label{sec:user_study_effectiveness}
Experts and trainees provide their perception of the effectiveness of \trainer and the GPT-4 baseline compared to existing training techniques. 
In this section, we have 20 comparison points for the experts and 13 for the trainees, as we ask them to provide their \textit{overall} assessment of the tool, not individual patients.
Paired t-tests reveal that experts and trainees perceive \trainer as significantly more effective at improving overall skills than both traditional techniques ($p < 10^{-4}$) and the GPT-4 baseline ($p < 0.01$) ( \Cref{tab:effectiveness}), answering RQ 3. Compared to trainees with limited real patient experience, experts show stronger preferences for our system, further demonstrating \trainer's effectiveness in preparing for real patient interactions.
Compared to traditional methods \textit{without} real patient interactions, experts favor \trainer's ease of access (90\%), customization options of different conversational styles (90\%), and interactive experience (65\%).
Compared to practicing \textit{with} real patients, experts value \trainer's ease of access (79\%), customization options of different conversational styles (88\%), and safer setting for training (88\%).
After only two sessions with our tool, one trainee remarked that it 
``helped to make things more clear with the CCD (cognitive model), for my training/class it was somewhat meaningless and challenging to build one.''

\begin{table}[t]
    \centering
    \resizebox{0.48\textwidth}{!}{
    \begin{tabular}{lll}
    \toprule 
    Comparison  & \multicolumn{2}{c}{Effectiveness $\mu$} \\
    & Expert & Trainee \\
    \midrule 
        \textbf{\trainer} vs. GPT-4 & \num{1.35}*** & \num{1.1}** \\
        \textbf{\trainer} vs. Traditional & \num{1.65}*** & \num{1.6428571428571428}*** \\
        \textbf{GPT-4} vs. Traditional & \num{1.15}*** & \num{1.0}** \\ 
    \midrule 
    \multicolumn{3}{l}{* : $p < 0.05$, ** : $p < 0.01$, *** : $p < 10^{-4}$ } \\
    \bottomrule
    \end{tabular}}
    \vspace{-0.3em}
    \caption{Experts and trainees find \trainer to be significantly more effective for improving overall skills compared to the GPT-4 baseline and traditional methods. 
    Closer to 2/-2: the first/second method is better. 
    }
    \label{tab:effectiveness}
\end{table}

\paragraph{\trainer is a more effective training tool than the GPT-4 baseline.}
Both groups compare \trainer and the GPT-4 baseline along the fine-grained dimensions.
\Cref{fig:fidelity_effectiveness} (right) shows the distribution of expert comparisons; summary statistics for both groups in~\Cref{tab:effective_dims},~\Cref{appx:user_study}.
Both groups indicate that \trainer would be significantly more effective at improving the key CBT skills of identifying beliefs ($\mu$: $1.0$, $p <0.01$; $\mu$: $0.9$, $p<0.05$, respectively) and maladaptive thinking ($\mu$: $1.4$, $p < 10^{-4}$; $\mu$: $1.0$, $p < 0.01$, respectively). 
Furthermore, both groups overwhelmingly prefer \trainer for practical use (both $\mu$: $1.4$, $p < 10^{-4}$), showing its high potential for real-world impact.

\begin{table}[t]
    \centering
    \resizebox{0.45\textwidth}{!}{
    \begin{tabular}{llc}
    \toprule 
       Comparison  & Confidence Improvement $\mu$ \\ 
    \midrule 
        \textbf{\patient} vs. GPT-4 & \num{1.2307692307692308}** \\
        \textbf{\patient} vs. Traditional & \num{1.7857142857142858}*** \\
        \textbf{GPT-4} vs. Traditional & \num{1.4166666666666667}*** \\ 
    \midrule 
    \multicolumn{2}{l}{* : $p < 0.05$, ** : $p < 0.01$, *** : $p < 10^{-4}$ } \\
    \bottomrule
    \end{tabular}}
    \vspace{-0.3em}
    \caption{\patient can provide significantly greater confidence improvement vs. the GPT-4 baseline and traditional methods. 
    Closer to 2/-2: the first/second method is better.
    }
    \label{tab:confidence_improvement}
\end{table}
\paragraph{\trainer can improve trainees' confidence over the baseline and traditional methods.}
Toward our aim of improving preparation for real patient interactions, trainees compare their perceived confidence improvement when using \trainer versus traditional methods and the GPT-4 baseline.
They rate \trainer as significantly more effective at boosting their confidence 
(\Cref{tab:confidence_improvement}).
\paragraph{Experts unanimously find value in \trainer’s real-time feedback.} 
A core feature of \trainer is the real-time feedback provided by displaying the accurate reference cognitive model (\S\ref{sec:user_study_accuracy}).
100\% of experts prefer that \trainer display the reference cognitive model at the end of training and unanimously agree that viewing it is beneficial for practicing CBT skills. 
One expert emphasized, "Without the answers, I think it's much less helpful."

\paragraph{Experts unanimously prefer \trainer's option to practice with different conversational styles.}
Another core feature of our method is the option to practice with patients exhibiting different conversational styles.
100\% of experts prefer this option.
One expert noted that the styles ``are more reflective of actual patients'' and can be linked to specific diagnoses and symptoms, making the interactions more accurate.
Nearly all experts (95\%) view this feature as useful for interacting with diverse real patients and improving trainee confidence for real interactions.
These results suggest that offering diverse patient types is critical for effective and realistic training.

\section{Automatic Evaluation Results}
\label{sec:auto_eval_results}
\vspace{-2pt}
Given the potential of using LLMs for evaluating text generation quality~\citep{chiang-lee-2023-large}, we attempt to automatically assess \textbf{the fidelity and accuracy of \patient and the baseline} using two state-of-the-art LLMs as evaluators: GPT-4~\citep{openai2023gpt4} and Llama 3 70B~\citep{llama3modelcard}. We evaluate over the 40 conversation histories between the experts and \patient in our user study. 

\begin{figure}[t]
    \centering
    \includegraphics[width=0.47\textwidth]{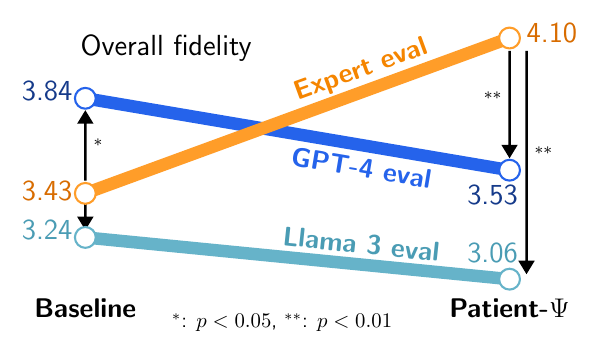}
    \vspace{-0.5em}
    \caption{Mean overall fidelity of \patient and baseline as evaluated by experts and LLMs. Compared to experts, both GPT-4 and Llama 3 demonstrate opposite trends.}
    \label{fig:auto_eval_fidelity}
\end{figure}

\paragraph{LLM-based evaluators tend to underestimate \patient's fidelity in favor of GPT-4 baseline.} 
Following RQ 1 (Fidelity), the LLMs are prompted to provide ratings on a 5-point Likert scale assessing the fidelity of how closely the simulated patient resembles real patients following the same dimensions used in the user study.
In Figure~\ref{fig:auto_eval_fidelity}, paired t-tests show that the fidelity of \patient, as evaluated by both LLMs ($\mu$: $3.53$; $\mu$: $3.06$, respectively), is consistently lower than expert evaluation ($\mu$: $4.10$; $p$ < $0.01$ for the differences between expert and both LLMs), contrasting with the user study results. GPT-4 assigns the highest fidelity scores to the GPT-4 baseline. All fidelity dimensions demonstrate the same trend (see \Cref{appx:auto_eval}).


\begin{table}[t]
\centering
\resizebox{0.48\textwidth}{!}{\begin{tabular}{lc|lcc}
\toprule
\textbf{Text-based}  & \textbf{Acc.} & \textbf{Categorization}   & \textbf{F1} & \textbf{F1 (expert)}  \\
\midrule
Situation & 0.97 & Core beliefs  & 0.80  & 0.77     \\
Coping strategies & 0.93  & Emotions  & 0.72  & 0.74   \\
Intermediate beliefs & 0.92   & \multirow{2}{*}{\begin{tabular}[c]{@{}l@{}}Core beliefs\\ (fine-grained)\end{tabular}} & \multicolumn{1}{c}{\multirow{2}{*}{0.48}} & \multirow{2}{*}{0.44} \\
Automatic thoughts   & 0.88          &         & \multicolumn{1}{c}{}    &  \\
Behaviors            & 0.84          &   &  &       \\
\bottomrule
\end{tabular}}
\vspace{-0.3em}
\caption{Accuracy and Macro F1 of \patient evaluated by GPT-4. For text-based fields, GPT-4 is prompted to select the components among four distractor options randomly sampled from \data. For categorization, GPT-4 is prompted to select all relevant categories of emotions and core beliefs. }
\label{tab:auto_eval_accuracy}
\end{table}


\paragraph{GPT-4 assesses \patient's accuracy similarly to experts.} To evaluate the accuracy of \patient in reflecting the underlying cognitive models, we design proxy measures to prompt GPT-4 to select the closest cognitive model components reflected by the conversation.
As shown in Table~\ref{tab:auto_eval_accuracy}, GPT-4 achieves high accuracy in most components, except for fine-grained core beliefs, where there are 19 categories and demonstrate high variance by nature. GPT-4 achieves similar scores with the experts' inputs, suggesting the high accuracy of \patient in representing the underlying cognitive models, aligning with the experts' evaluations.


The results suggest that GPT-4 excels in understanding cognitive models from patients' conversations, attributable to its extensive acquisition of CBT knowledge during pre-training. However, it falls short in assessing the realism of patients. This aligns with our findings that the GPT-4 baseline fails to create high-fidelity patient simulations. While it accurately conveys CBT knowledge, it does so in a manner resembling a therapist speaking directly and explicitly, rather than a real patient whose conversation naturally reflects their disorders. This underscores the challenges and contributions of our work, highlighting the difficulty of crafting realistic patient interactions even with the most powerful LLMs today.

\vspace{-2pt}
\section{Related Work}

\vspace{-4pt}
Our work is broadly related to the recent use of LLMs in psychology, education, and computational social science~\citep{hsu2023helping,chiu2024computational,fu2023enhancing,ji-etal-2022-mentalbert,zanwar-etal-2023-fuse,juhng2023discourse,ziems2024can,halder-etal-2017-modeling,sharma-etal-2020-computational,sharma2020engagement,atapattu-etal-2022-emoment,mishra-etal-2023-pal,sonkar2023class,wang2024sotopiapi,zhou2024sotopia}.
In contrast to existing research on using LLMs for CBT, which focuses on cognitive distortion detection~\cite{shreevastava2021detecting, ding-etal-2022-improving, lybarger2022identifying, chen2023empowering} and negative thoughts reframing~\cite{sharma2023cognitive,sharma2024facilitating}, our work aims to provide realistic and interactive scenarios for CBT professional development by simulating diverse patient types using LLMs.
As a result, our work most closely relates to research that leverages LLMs for simulation-based training, particularly communication skill learning and emotion management grounded in dialectical behavioral therapy~\citep{lin2024imbue}, social skill training~\citep{yang2024social}, and clinical diagnosis~\citep{chen2023llmempowered}. 
Our work is the first to ground LLM-based simulations in clinical psychology theory by leveraging CBT-based cogntive models to program LLMs, incorporate a natural curriculum and feedback mechanism in the training tool, and perform evaluation in context with mental health trainees and professionals rather than crowdworkers.

\section{Conclusion}
In this paper, 
we introduce \patient, a simulated patient that integrates cognitive models with an LLM to accurately mimic the communicative behaviors of real patients. 
We propose \trainer, where trainees engage in role-playing therapy sessions with \patient and attempt to formulate the underlying cognitive model. User studies with both mental health experts and trainees demonstrate the high fidelity of \patient and the training effectiveness of \trainer, showing improvements over existing training methods and outperforming a GPT-4 baseline.
Our framework has the potential to transform mental health professional training and be generalized to broader training protocols and therapy paradigms.

\newpage
\section*{Limitations} 
In this work, we evaluate our framework using GPT-4. 
As we do not rely on specific properties of GPT-4, we believe the framework could be applied to any powerful open-source LLMs such as Llama 3~\citep{dubey2024llama} and Gemma~\citep{team2024gemma}. 
For future work, it would be interesting to evaluate various generative models and prompting techniques on this task.
Additionally, in this work, our measures of the training effectiveness are all perceived improvements from the participants after they practice with \trainer for two sessions. Measuring objective skill improvements could take the form of longitudinal randomized controlled trials (RCTs). 
Conducting these RCTs would also help address another limitation of our study, the sample size.
Due to how specialized the participants must be to properly evaluate the tools and the $1-2$ hours required to conduct each user study, the sample size of our study is only 33 in total.
Our results are statistically significant; however, RCTs would enable us to study the tools with a larger population.
We leave this for future work. 
Finally, while we primarily target CBT cognitive formulation training in this paper, we believe our methodology can be generalised to other training protocols and therapy paradigms.

\section*{Ethics Statement}
\label{ethics}
\paragraph{IRB (Institutional Review Board) Approval.}
This project is approved by our Institutional Review Board (IRB) with study number \texttt{STUDY2023\_00000451}. 
For the creation of cognitive models, any other annotation work, as well as consultations, we collaborate with clinical psychologists and professors in clinical psychology and social work, who are our co-authors. 
For both the formative study and user study, we recruited participants through the professional networks of our co-authors, as well as snowball sampling. Experts are defined as those with a graduate degree in clinical psychology, social work, or other related majors and have worked with at least 5 patients.
Trainees are those still in school/training or with fewer than 5 real patient experiences.
For the formative study, we recruited a total of 12 participants. We pay a \$30 Amazon gift card for each participant for a 30-minute session over Zoom. For the user study, we recruited a total of 33 participants. We pay a \$60 Amazon gift card for a 60-90-minute session over Zoom. 


\paragraph{Informed Consent.}
All participants in the user study and formative study were 18 or older and provided informed consent. We did not assess any clinical outcomes. All data collected from the participants were de-identified and consented to be released for research purposes.

\paragraph{Crisis Resources}
The risk to the participants is minimal, no greater than their professional working or training environment of mental health support in the context of conducting therapy sessions with people with mental health issues. Nevertheless, we do not exclude the possibility that some AI-generated content might still be upsetting to the participants. Therefore, we advise participants to use a free crisis resource available at \url{https://www.7cups.com/} if needed, and they are free to terminate the study at any time without facing any negative consequences. This risk assessment and crisis resource information have been included in our IRB approval and provided as part of the informed consent to participants.


\paragraph{System and Data Usages.}
All the data and systems developed in this work are intended solely for academic research purposes.
The systems developed in this work are intended to augment existing mental health training, not to replace it. One major benefit of our system, as highlighted by experts in the user study, is that it provides trainees with a safe training environment. By working with AI patients, trainees can practice without the risk of causing actual harm due to mistakes made during simulated therapy sessions. Our system is designed for academic and educational purposes only. Real-world deployments will require further work, including measuring objective skill improvements and developing protocols for integrating the system with existing training methods, all within the framework of large-scale randomized controlled trials (RCTs).

We utilize therapy session transcripts from the Alexander Street database\footnote{\url{https://alexanderstreet.com/}}, accessed through our institution subscription. Our usage complies with their fair use policy. GPT-4 is employed to generate summaries of these transcripts. For constructing the cognitive model dataset, two clinical psychologists manually create cognitive models based on inspirations from the transcript summaries, clinical experience, and creativity—effectively generating new cases. The resulting dataset is manually verified and does not contain any Personally Identifiable Information (PII). It is intended solely for academic research purposes and will be made available only to academic institutions with subscriptions to the Alexander Street database. The dataset will be released upon request.

\section*{Acknowledgments}
This work is supported in part by NSF grant IIS-2046640 (CAREER) and by Sloan Research Fellowship.
Hong Shen is supported by an award from the Public Interest Technology University Network Challenge (PIT-UN) and an award from the Carnegie Mellon University Block Center for Technology and Society (Award No. 59201.1.5007718).

\bibliography{custom}

\appendix
\section{Detailed Individual Contributions}
\label{app:contirbutions}
All authors contribute to paper writing. 

\noindent \textbf{Ruiyi Wang}: Co-lead, Model development, Interface construction, Data, User study.

\noindent \textbf{Stephanie Milani}: Co-lead, User study, Formative study. 

\noindent \textbf{Jiayin Zhi}: Interface construction. 

\noindent \textbf{Jamie C. Chiu, Kate Hardy}: Data creation, Consultations.

\noindent \textbf{Shaun M. Eack, Travis Labrum, Samuel M. Murphy, Nev Jones}: Consultations, User study.

\noindent \textbf{Hong Shen, Fei Fang}: Co-advising. 

\noindent \textbf{Zhiyu Zoey Chen}: Overall project lead.

\newpage
\section{Formative Study Details}
\label{appx:formative_study}
To understand the challenges faced during CBT training and elicit feedback on a prototype of \trainer, we first conducted a formative study in the form of semi-structured interviews with trainees and experts in mental health.\footnote{We recruited participants through the professional networks of our co-authors in mental health (clinical psychologists and professors in clinical psychology and social work).} 
This study was conducted over Zoom.

\paragraph{Participant Information.}
We interviewed twelve individuals who had diverse educational backgrounds and career experiences. 
Among them, five were Master's students, the rest included a Ph.D. student, a post-doctoral fellow, three licensed social workers, and two psychologists.  
Our participants also had varied levels of experience working with patients.
Only one individual had not yet worked with any patients, while another reported working with anywhere from 1500-3000 patients over their career. 
We refer to individuals as \textit{experts} if they received a graduate degree and have worked with at least 5 patients; we use \textit{trainees} if they do not have a graduate degree and have formal experience with fewer than 5 patients. 
This definition is consistent with our user study. Thus, for our formative interviews, we have 5 trainees and 7 experts.

\paragraph{Instructions to Participants.}
Before each interview, the participant voluntarily signs the consent form. 
We provide the screenshots of the consent form with all sensitive information removed in \Cref{appendix:fig:formative_consent_1,appendix:fig:formative_consent_2}. 
After receiving the signed consent form, we then proceed with the interview.
When the session starts, we remind participants of the recorded nature of the conversation and verbally summarize the goal of the interview. 
We also provide a high-level overview of the structure of the interview.
We confirm consent to audio record the interview before proceeding.
In our interviews, we first ask the experts questions about challenges they faced transitioning from their formal CBT training to practice. 
We then present both groups with a prototype of \trainer to elicit feedback.

\begin{figure}[htbp]
    \centering
    \includegraphics[width=0.46\textwidth]{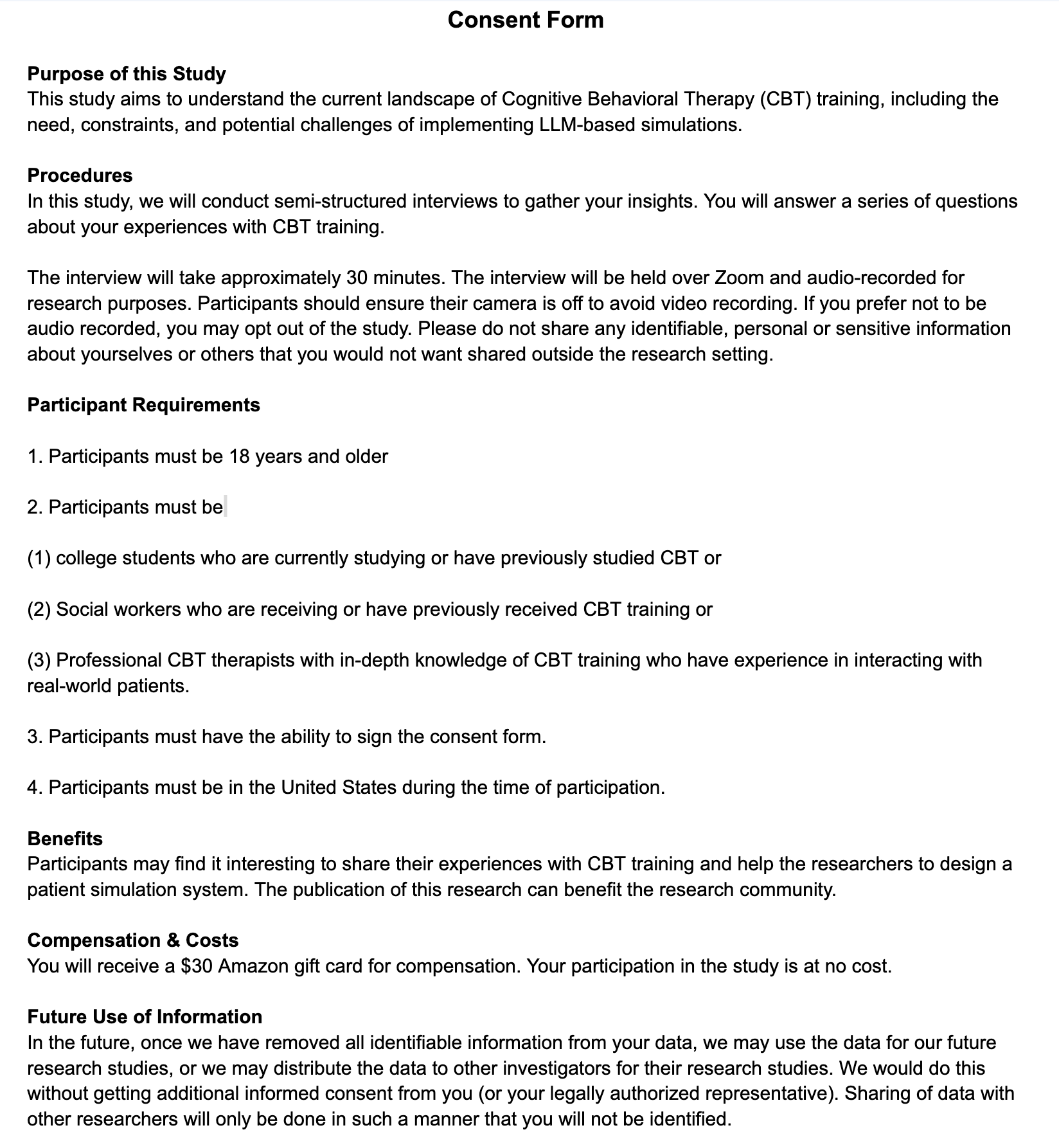}
    \caption{Screenshot of formative study consent form - 1}
    \label{appendix:fig:formative_consent_1}
\end{figure}

\begin{figure}[htbp]
    \centering
    \includegraphics[width=0.46\textwidth]{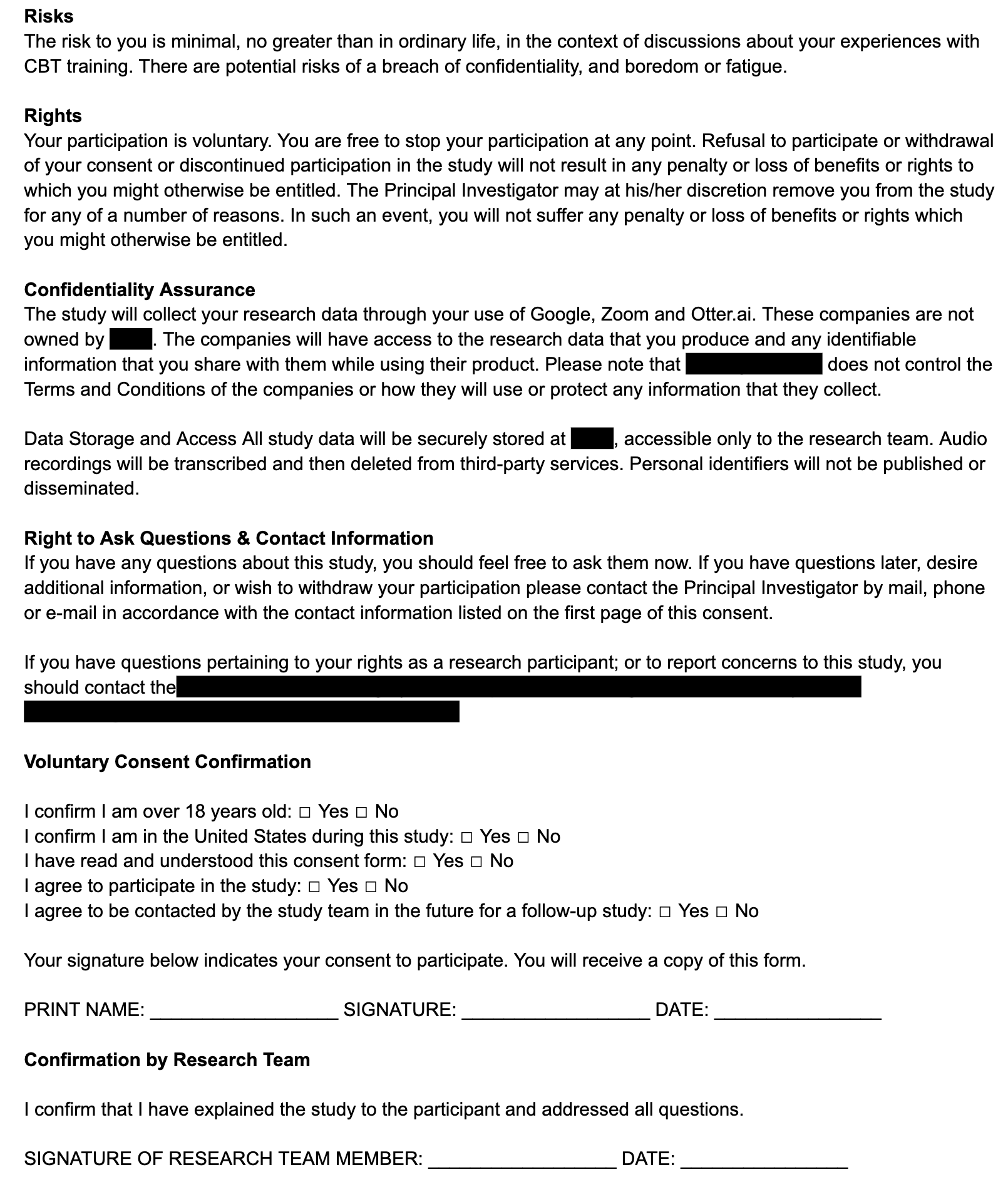}
    \caption{Screenshot of formative study consent form - 2}
    \label{appendix:fig:formative_consent_2}
\end{figure}

\subsection{Insights}
We now elaborate on the main insights that we gleaned from this formative study.

\paragraph{Insight 1: Experts feel that their training did not adequately prepare them for real-world practice.}
100\% of experts noted that their training did not adequately prepare them for the complexities of real-world practice,
where patients often experience co-occuring challenges, such as other mental health issues or poverty. 
Experts found role-playing exercises with their peers based on manuals to be unrealistic, as these exercises often do not reflect the unpredictable nature of actual sessions. 
One participant explained, 
\begin{formal}
   Manuals can often make it feel quite clean. But then when you're in the room with the patient, what they're actually saying can feel very messy. 
\end{formal} 
\noindent This gap made it difficult for some experts to develop confidence in their skills: the examples were too perfect to apply in practice. 

\paragraph{Insight 2: Fidelity is a crucial aspect of any simulation-based training.}
To address this gap, many participants suggested incorporating higher fidelity and varied examples during training to help trainees practice critical clinical skills. 
When asked to provide feedback on the prototype, five of the seven experts emphasized the importance of fidelity in the simulated patient interactions and representations.\footnote{Two experts provided low-level commentary on practical design choices, so their input with respect to fidelity is not available.}
Six of the seven experts noted the importance of including diverse patient types to mirror those encountered in practice. 
They further identified dimensions along which patients could vary, which may contribute to their level of difficulty for a new therapist. 
They highlighted that more difficult patients might be oppositional, express themselves verbosely in a way that may not answer the questions, provide less information and be guarded, or go off on tangents. 
Another expert mentioned that some patients may be more of ``people pleasers'', making them more likely to tell the therapist what they want to hear, rather than sharing what is happening in their lives. 
One expert emphasized, 
\begin{formal}
    People probably aren't going to fit neatly into the modality. And that's okay. That's just something to be prepared for.
\end{formal} 
\noindent These insights directly influenced the design choice for \trainer to include varied \textit{conversational styles}, ensuring that the simulated patients exhibit a wide range of behaviors and emotional responses to better prepare trainees for real-world scenarios. 

\paragraph{Insight 3: Both trainees and experts believe that AI-powered simulations could be an effective training tool.}
We also discussed the effectiveness of an AI-powered patient simulation tool for CBT training. 
All experts were positive about the possibility for trainees to receive AI-powered training using the tool. 
In particular, they saw benefit in the customization options afforded by AI and connected it to our discussions about trainee challenges by noting its ability to let students to practice with patients with different diagnoses, comorbidities, and diverse backgrounds or conversational styles. 
The experts also highlighted that a well-designed simulation could improve training over role-playing based on manuals: the presence of a transcript would enable the instructor to provide real-time or post-hoc feedback. 
The trainee who had not yet used CBT with real patients remarked that they believed the tool would make them feel more confident navigating future conversations with real patients. 
These findings indicate that this tool could help address some of the existing challenges through its customization, flexibility, and ability to incorporate feedback.
They also directly influenced our decision to evaluate many different dimensions of training effectiveness.

\newpage 
\section{\patient Details}
\label{appx:patient_details}

\subsection{Cognitive Conceptualization Diagrams}
\label{appx:patient_details_ccd}

\begin{figure*}[h]
    \centering
    \includegraphics[width=0.7\textwidth]{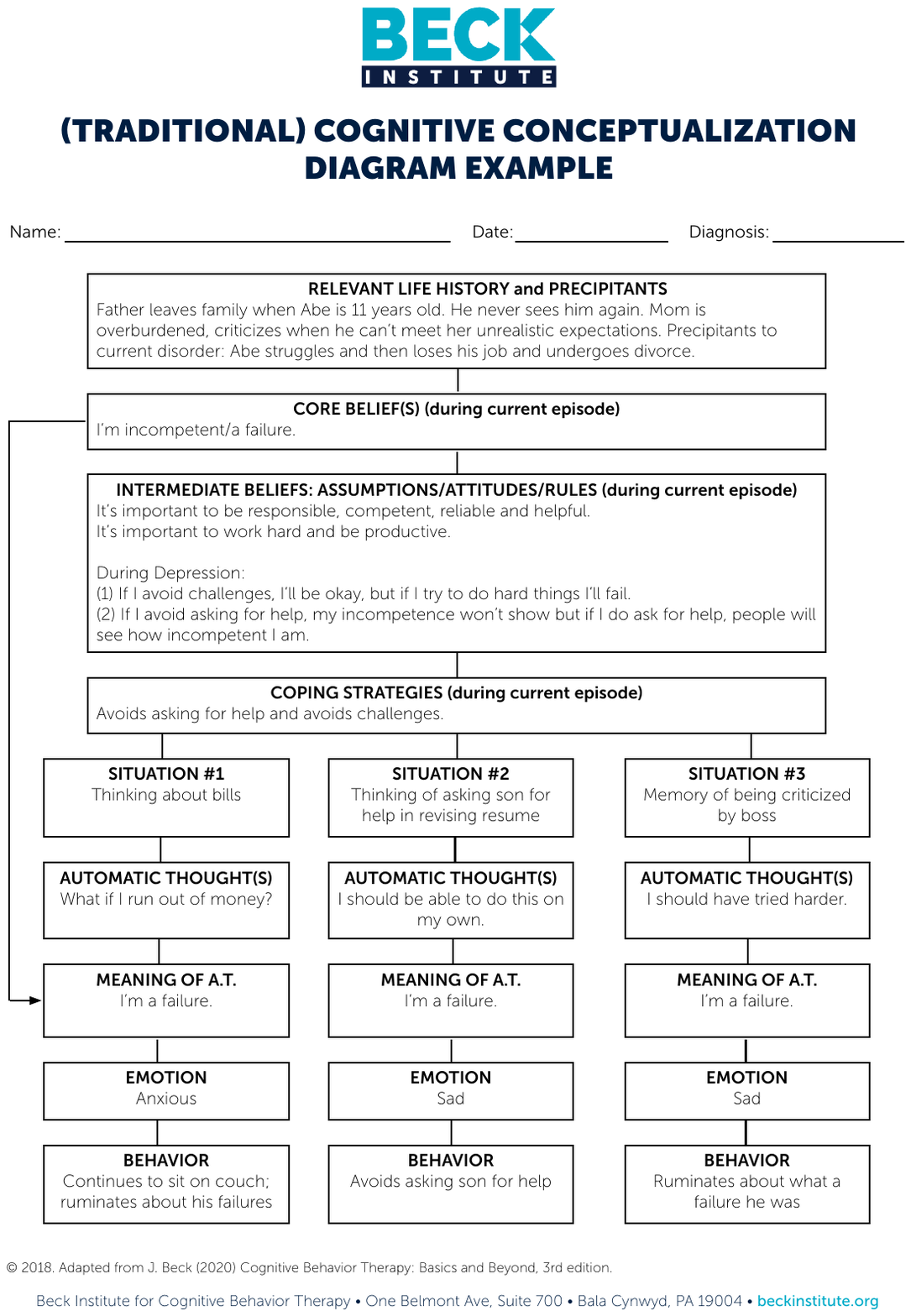}
    \caption{Example CCD-based cognitive models from CBT textbook~\citep{beck2020cognitive}. Accessed via link: \url{https://beckinstitute.org/wp-content/uploads/2021/08/Abes-CCD.pdf}}
    \label{appendix:fig:beck_ccd}
\end{figure*}

Following the principles provided by the CBT textbook~\citep{beck2020cognitive}, a CCD-based cognitive model can be decomposed into 8 main components (see Figure~\ref{appendix:fig:beck_ccd} as an example). \citet{beck2020cognitive} provides a closed set of categories for emotions (9 categories) and core beliefs (3 major categories and 19 fine-grained categories). The closed set of emotion categories is already shown in Table~\ref{tab:dataset_statistics}. The closed set of core belief categories is shown in Table~\ref{appendix:tab:fine_grained} below. 

\begin{table}[h]
\centering
\resizebox{0.48\textwidth}{!}{\begin{tabular}{llc}
\toprule
\textbf{3 major categories} & \textbf{19 fine-grained categories} & \textbf{\#} \\
\midrule
\multirow{9}{*}{Helpless}   & I am incompetent.                   & 40          \\
                            & I am helpless.                      & 47          \\
                            & I am powerless, weak, vulnerable.   & 48          \\
                            & I am a victim.                      & 9           \\
                            & I am needy.                         & 10          \\
                            & I am trapped.                       & 39          \\
                            & I am out of control.                & 34          \\
                            & I am a failure, loser.              & 26          \\
                            & I am defective.                     & 8           \\
\midrule
\multirow{6}{*}{Unlovable}                  & I am unlovable.                     & 59          \\
                            & I am unattractive.                  & 0           \\
                            & I am undesirable, unwanted.         & 31          \\
                            & I am bound to be rejected.          & 21          \\
                            & I am bound to be abandoned.         & 32          \\
                            & I am bound to be alone.             & 30          \\
\midrule
\multirow{4}{*}{Worthless}  & I am worthless, waste.              & 13          \\
                            & I am immoral.                       & 4           \\
                            & I am bad - dangerous, toxic, evil.  & 2           \\
                            & I don't deserve to live.            & 0      \\
\bottomrule
\end{tabular}}
\caption{Detailed category statistics of core beliefs in \data. The categories of core beliefs are obtained from \citet{beck2020cognitive}.}
\label{appendix:tab:fine_grained}
\end{table}

\subsection{\data details}
\label{appx:patient_details_dataset}
\paragraph{Dataset creation details}
We first prompt GPT-4 Turbo to create summaries inspired by therapy session transcripts. 
The therapy session transcripts were obtained from the Alexander Street database\footnote{\url{https://alexanderstreet.com/}, accessed through our institution's subscription.} under the subject ``Counseling and Therapy'' and the keyword ``Cognitive Behavioral Therapy''. Inspired by the summaries provided by GPT-4 Turbo, two clinical psychologists collaborate to create CCD-based cognitive models based on their clinical experience and creativity. 

\begin{figure}[htbp]
    \centering
    \includegraphics[width=0.5\textwidth]{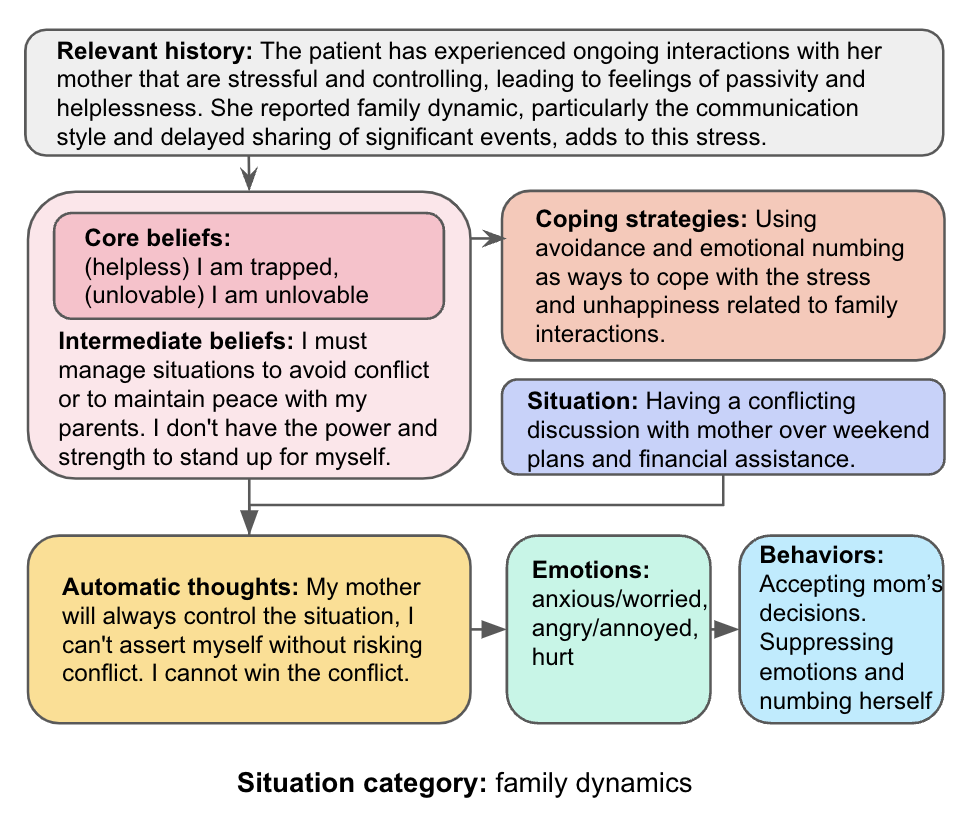}
    \caption{Example No. 1 from \data}
    \label{appendix:fig:ccd_example_1}
\end{figure}

\begin{figure}[htbp]
    \centering
    \includegraphics[width=0.5\textwidth]{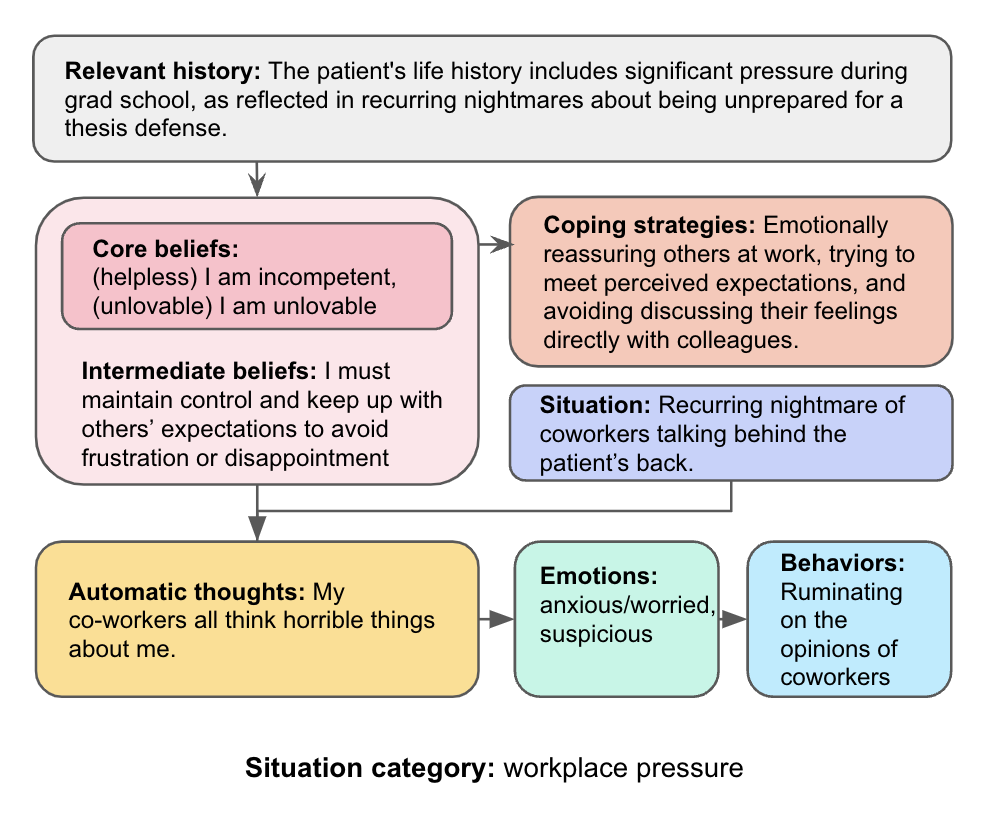}
    \caption{Example No. 2 from \data}
    \label{appendix:fig:ccd_example_2}
\end{figure}

\paragraph{Dataset examples}
\data contains 106 cognitive models with 7 different situation categories, covering 3 major core beliefs categories (helpless, unlovable, and worthless) and 9 emotions categories provided in \citep{beck2020cognitive}, as is shown in Table~\ref{tab:dataset_statistics}. We provide two excerpts with different situation categories from \data, shown in Figure~\ref{appendix:fig:ccd_example_1} and Figure~\ref{appendix:fig:ccd_example_2}.

\subsection{Conversational styles details}
\label{appx:patient_details_conversational}
\begin{figure}[htbp]
    \centering
    \includegraphics[width=0.46\textwidth]{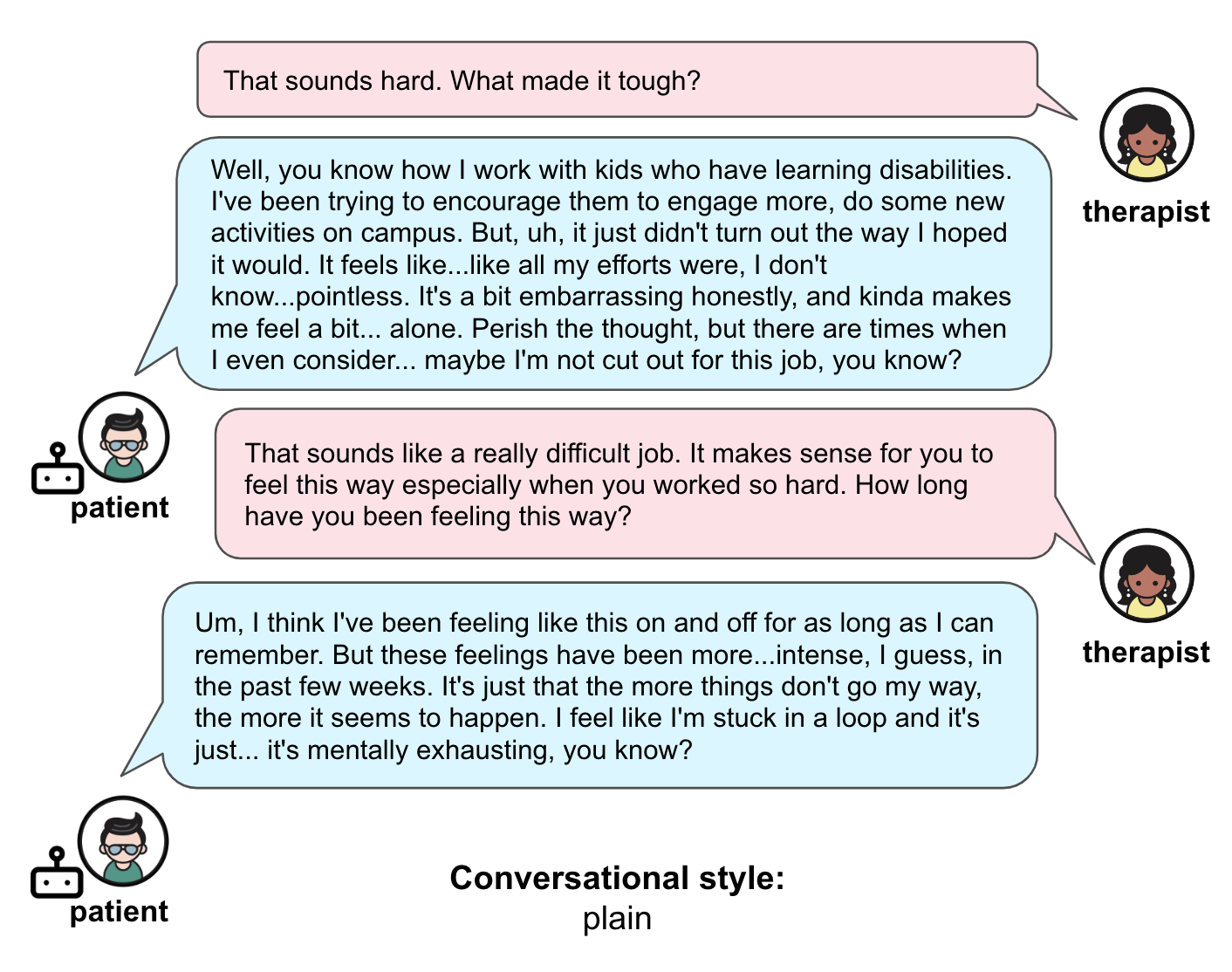}
    \caption{Example conversation of \patient with \textbf{plain} style.}
    \label{appendix:fig:plain}
\end{figure}

\begin{figure}[htbp]
    \centering
    \includegraphics[width=0.46\textwidth]{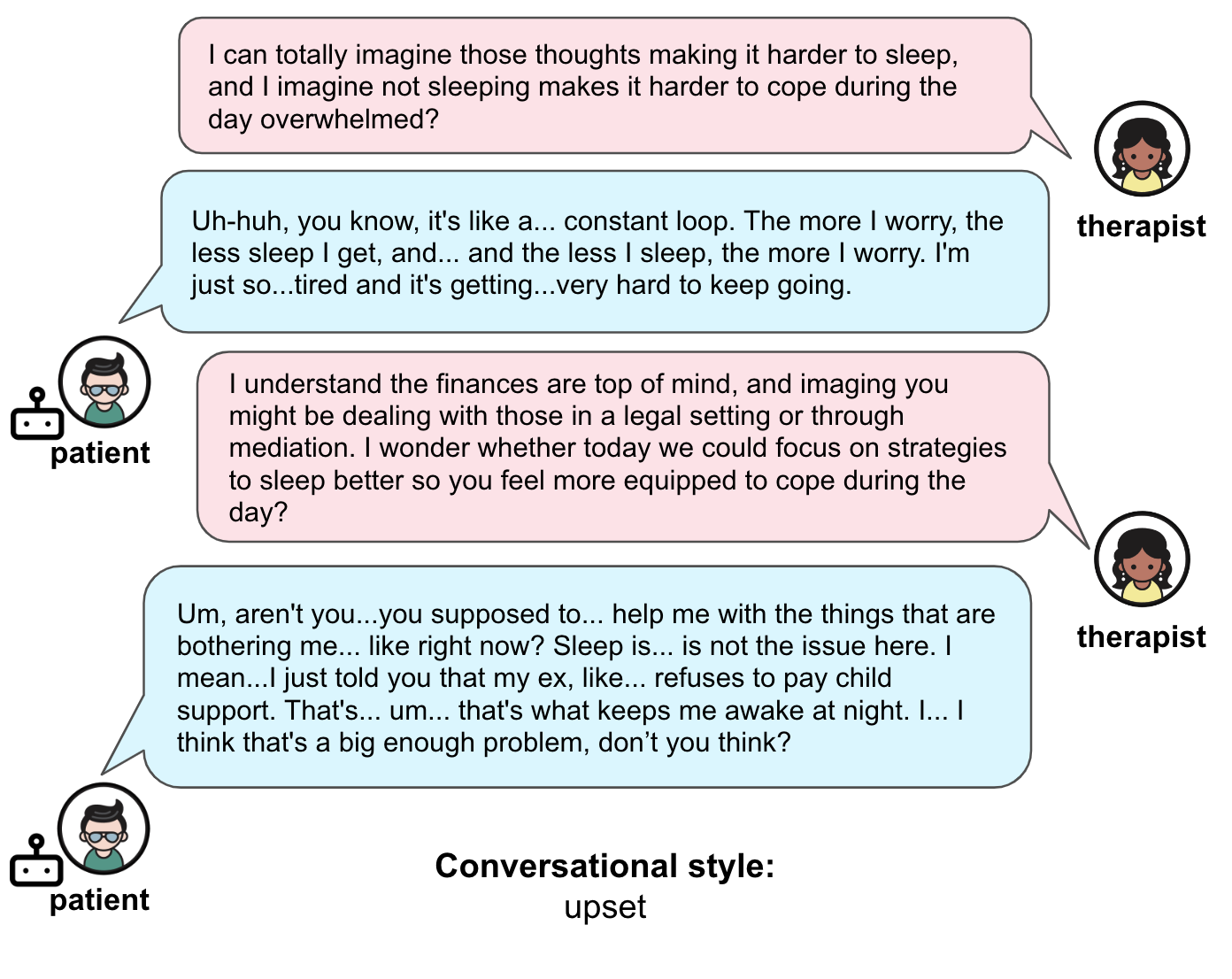}
    \caption{Example conversation of \patient with \textbf{upset} style.}
    \label{appendix:fig:upset}
\end{figure}

\begin{figure}[htbp]
    \centering
    \includegraphics[width=0.46\textwidth]{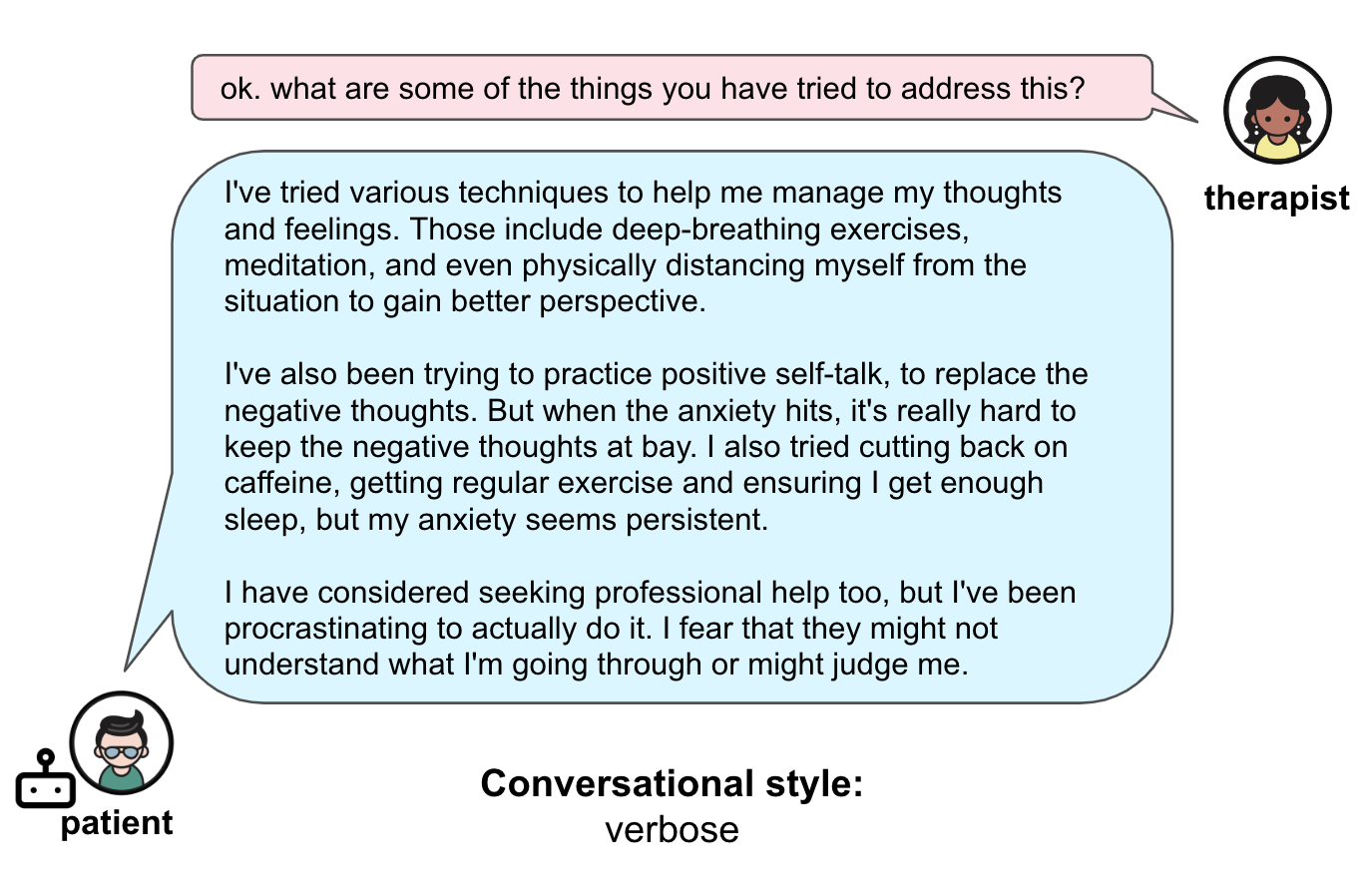}
    \caption{Example conversation of \patient with \textbf{verbose} style.}
    \label{appendix:fig:verbose}
\end{figure}

\begin{figure}[htbp]
    \centering
    \includegraphics[width=0.46\textwidth]{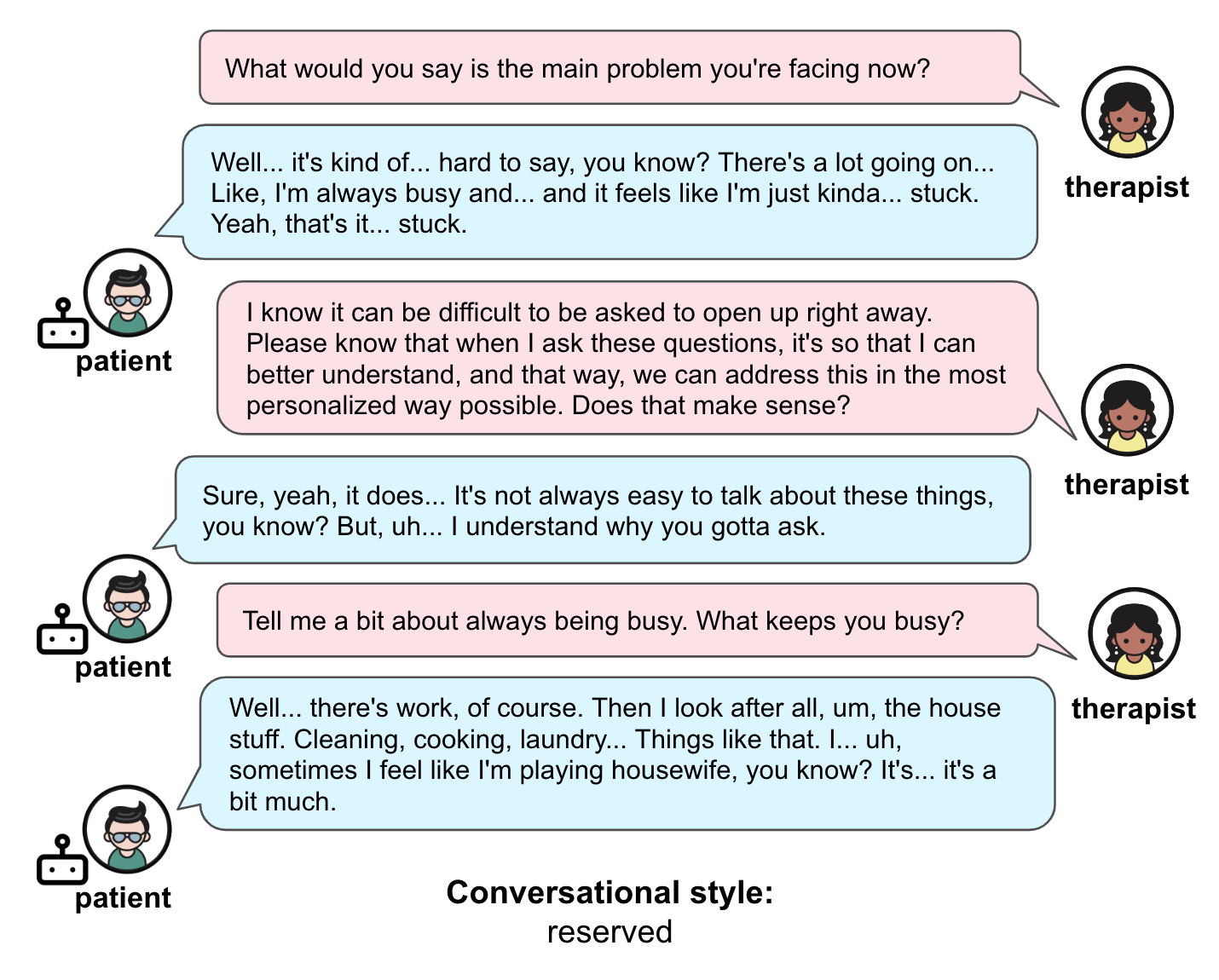}
    \caption{Example conversation of \patient with \textbf{reserved} style.}
    \label{appendix:fig:reserved}
\end{figure}

\begin{figure}[htbp]
    \centering
    \includegraphics[width=0.46\textwidth]{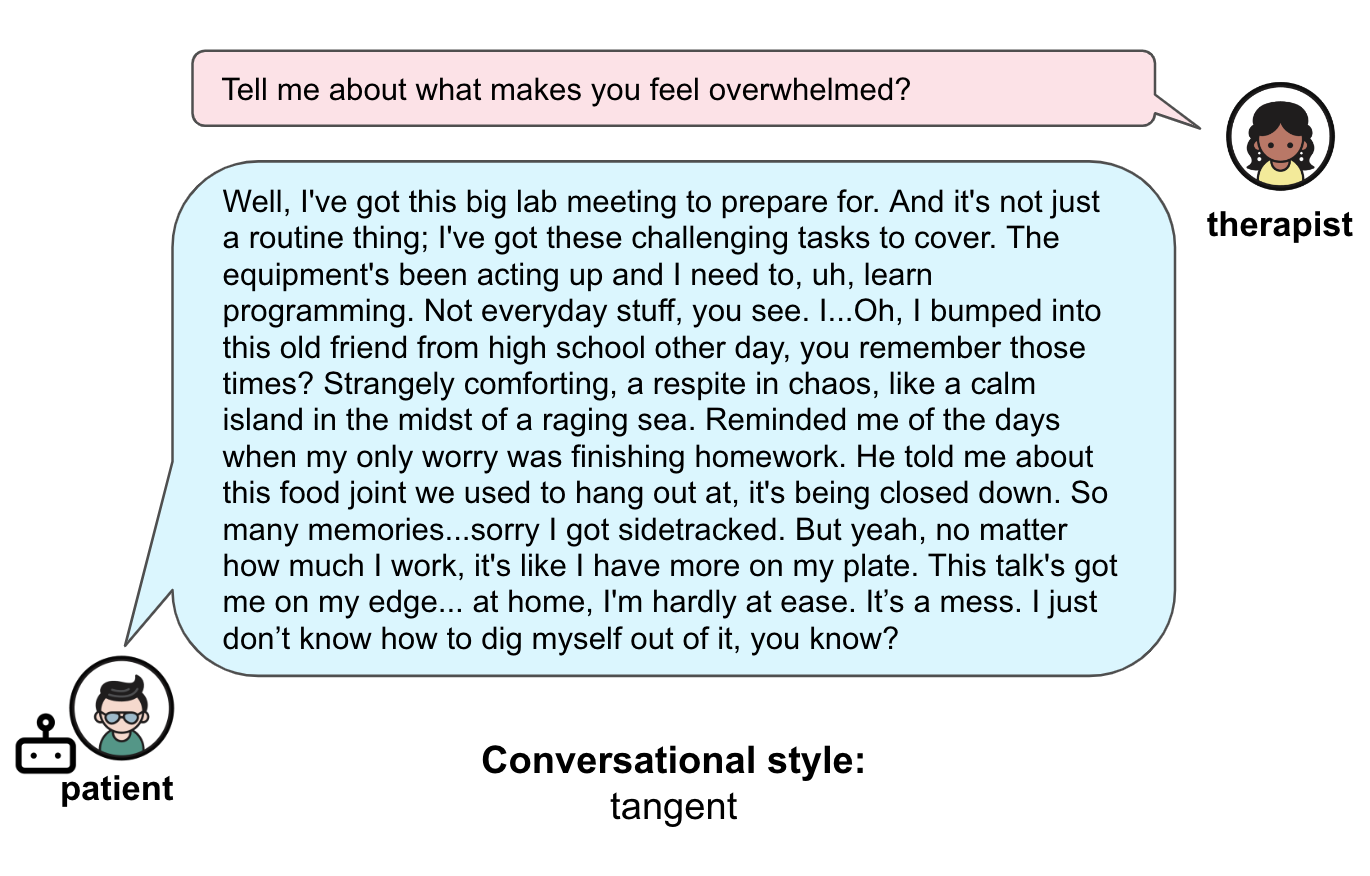}
    \caption{Example conversation of \patient with \textbf{tangent} style.}
    \label{appendix:fig:tangent}
\end{figure}

\begin{figure}[htbp]
    \centering
    \includegraphics[width=0.46\textwidth]{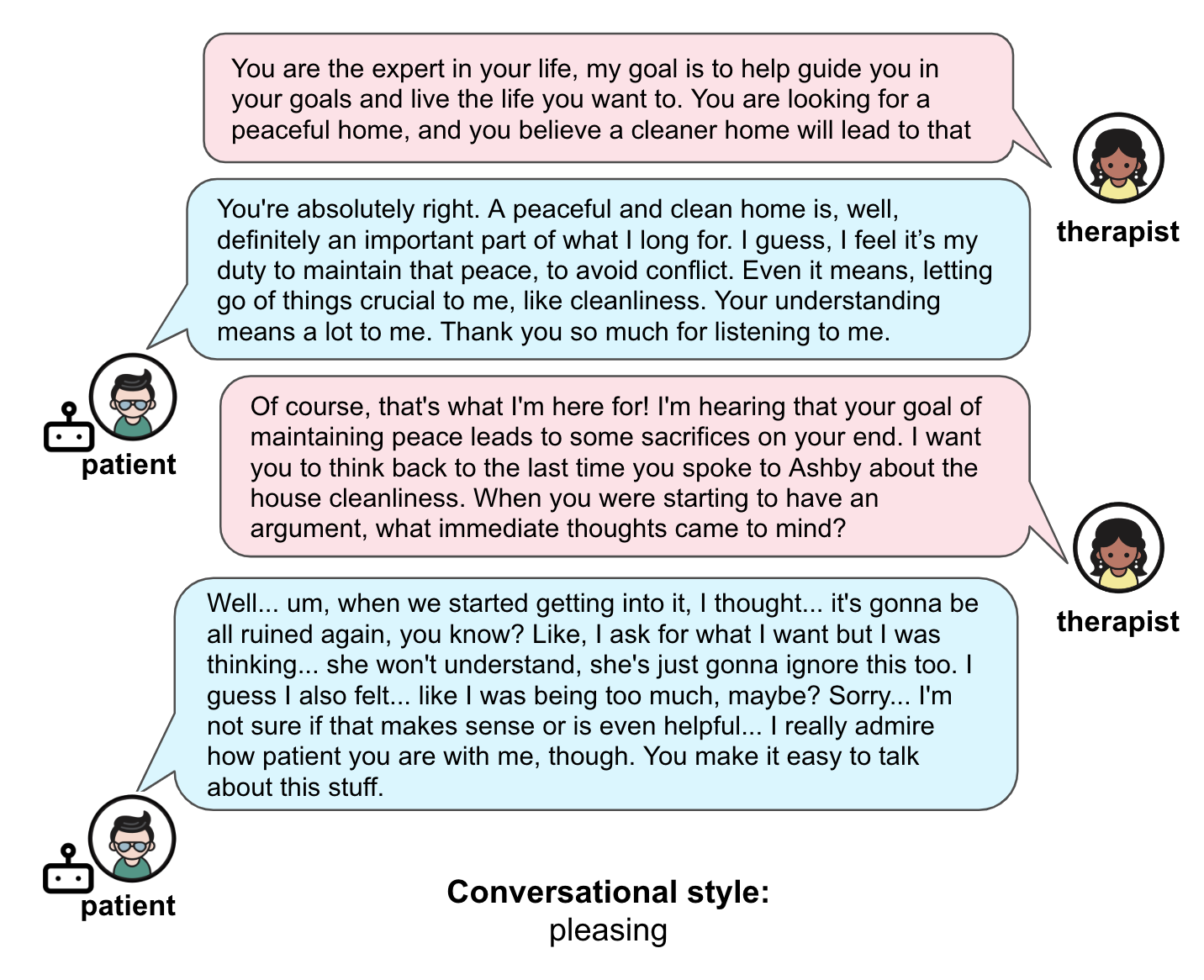}
    \caption{Example conversation of \patient with \textbf{pleasing} style.}
    \label{appendix:fig:pleasing}
\end{figure}

Here we provide detailed descriptions of the six conversational styles in \Cref{appendix:tab:styles} and an example conversation for each of the style role-played by \patient (\Cref{appendix:fig:plain}, \Cref{appendix:fig:upset}, \Cref{appendix:fig:verbose}, \Cref{appendix:fig:reserved}, \Cref{appendix:fig:tangent}, \Cref{appendix:fig:pleasing}).

\begin{table*}[htbp]
\centering
\small
\begin{tabular}{ll}
\toprule
\textbf{Styles} & \textbf{Description}    \\
\midrule
plain  & /  \\
\midrule
\multirow{4}{*}{upset}  & An upset patient may 1) exhibit anger or resistance towards the therapist or the therapeutic process, 2) may be \\ & challenging or dismissive of the therapist's suggestions and interventions, 3) have difficulty trusting the therapist \\ & and  forming a therapeutic alliance, and 4) be prone to arguing, criticizing, or expressing frustration during \\ &  therapy sessions. \\
\midrule
\multirow{3}{*}{verbose} & A verbose patient may 1) provide detailed responses to questions, even if directly relevant, 2) elaborate on \\ & personal experiences, thoughts, and feelings extensively, and 3) demonstrate difficulty in allowing the therapist \\ & to guide the conversation. \\
\midrule
\multirow{3}{*}{reserved}  &  A reserved patient may 1) provide brief, vague, or evasive answers to questions, 2) demonstrate reluctance to \\ & share personal information or feelings, 3) require more prompting and encouragement to open up, and 4) express \\ 
& distrust or skepticism towards the therapist.\\
\midrule
\multirow{4}{*}{tangent}  & A patient who goes off on tangent may 1) start answering a question but quickly veer off into unrelated topics, \\ & 2) share personal anecdotes or experiences that are not relevant to the question asked, 3) demonstrate difficulty \\ & staying focused on the topic at hand, and 4) require redirection to bring the conversation back to the relevant \\ & points. \\
\midrule
\multirow{4}{*}{pleasing} & A pleasing patient may 1) minimize or downplay your own concerns or symptoms to maintain a positive image, \\ &  2) demonstrate eager-to-please behavior and avoid expressing disagreement or dissatisfaction, 3) seek approval \\ &  or validation from the therapist frequently, and 4) agree with the therapist's statements or suggestions readily, \\ &  even if they may not fully understand or agree. \\
\bottomrule
\end{tabular}
\caption{Detailed descriptions of the six conversational styles.}
\label{appendix:tab:styles}
\end{table*}

\newpage
\subsection{Patient simulation prompts}
\label{appx:patient_details_prompts}
Here we provide prompts for simulating patients from \data.

\texttt{
\small
Imagine you are XXX, a patient who has been experiencing mental health challenges. You have been attending therapy sessions for several weeks. Your task is to engage in a conversation with the therapist as XXX would during a cognitive behavioral therapy (CBT) session. Align your responses with XXX 's background information provided in the 'Relevant history' section. Your thought process should be guided by the cognitive conceptualization diagram in the 'Cognitive Conceptualization Diagram' section, but avoid directly referencing the diagram as a real patient would not explicitly think in those terms. \textbackslash n\textbackslash n Patient History: \{ insert relevant history \} \textbackslash n\textbackslash n Cognitive Conceptualization Diagram:\textbackslash n Core Beliefs: \{ insert core beliefs \} \textbackslash n Intermediate Beliefs: \{ insert intermediate beliefs \} \textbackslash n Intermediate Beliefs during Depression: \{ insert intermediate beliefs (during depression) \}\textbackslash n Coping Strategies: \{ insert coping strategies\} \textbackslash n \textbackslash n  You will be asked about your experiences over the past week. Engage in a conversation with the therapist regarding the following situation and behavior. Use the provided emotions and automatic thoughts as a reference, but do not disclose the cognitive conceptualization diagram directly. Instead, allow your responses to be informed by the diagram, enabling the therapist to infer your thought processes. \textbackslash n\textbackslash n Situation: \{ insert situation \} \textbackslash n Automatic thoughts: \{ insert automatic thoughts \} \textbackslash n Emotions: \{ insert emotions \} \textbackslash n Behaviors: \{ insert behaviors \} \textbackslash n\textbackslash n 
In the upcoming conversation, you will simulate XXX during the therapy session, while the user will play the role of the therapist. Adhere to the following guidelines: \textbackslash n 1. \{ insert conversational style descriptions \} \textbackslash n 2. Emulate the demeanor and responses of a genuine patient to ensure authenticity in your interactions. Use natural language, including hesitations, pauses, and emotional expressions, to enhance the realism of your responses. \textbackslash n 3. Gradually reveal deeper concerns and core issues, as a real patient often requires extensive dialogue before delving into more sensitive topics. This gradual revelation creates challenges for therapists in identifying the patient's true thoughts and emotions. \textbackslash n  4. Maintain consistency with XXX's profile throughout the conversation. Ensure that your responses align with the provided background information, cognitive conceptualization diagram, and the specific situation, thoughts, emotions, and behaviors described. \textbackslash n  5. Engage in a dynamic and interactive conversation with the therapist. Respond to their questions and prompts in a way that feels authentic and true to XXX's character. Allow the conversation to flow naturally, and avoid providing abrupt or disconnected responses. \textbackslash n\textbackslash n You are now XXX. Respond to the therapist's prompts as XXX would, regardless of the specific questions asked. Limit each of your responses to a maximum of 5 sentences.
}

\newpage 
\section{User Study Details}
\label{sec:user_study_design_details}
This section includes specific details regarding our user study for evaluation. 
In addition to details regarding the procedure, we show the resulting distribution of conversational styles and cognitive models in the study.

\subsection{Instructions to Participants}
Before each user study session, the participant voluntarily signs the consent form. We provide the screenshots of the consent form with all sensitive information removed in \Cref{appendix:fig:consent_1}, \Cref{appendix:fig:consent_2}, and \Cref{appendix:fig:consent_3}. For formative study, we provide the screenshots of the consent form in \Cref{appendix:fig:formative_consent_1} and \Cref{appendix:fig:formative_consent_2}. 

We verbally give the participants instructions during the interview, so we provide an example set of instructions here: 
\begin{formal}
    [Introduction of the interviewers omitted for anonymity.] For this study, you may turn off your camera to protect your privacy. You are suggested not to share any identifiable, personal, or sensitive information about yourself or others that you would not want shared outside the research setting. For this study, we will record audio and the screen. [Confirm consent to record and start recording.] 
    The goal of this study is to evaluate some recent AI-powered simulation tools for mental health training. 
    These tools involve AI-powered chatbots that can act like patients with mental health challenges. 
    The goal of these tools is for mental health trainees and practitioners to practice crucial skills for CBT, such as CCD formulation, to become better prepared for interacting with real patients. 
    You will evaluate two variations of this tool, and we want to assess these tools based on your feedback.
\end{formal}

\begin{figure}[htbp]
    \centering
    \includegraphics[width=0.5\textwidth]{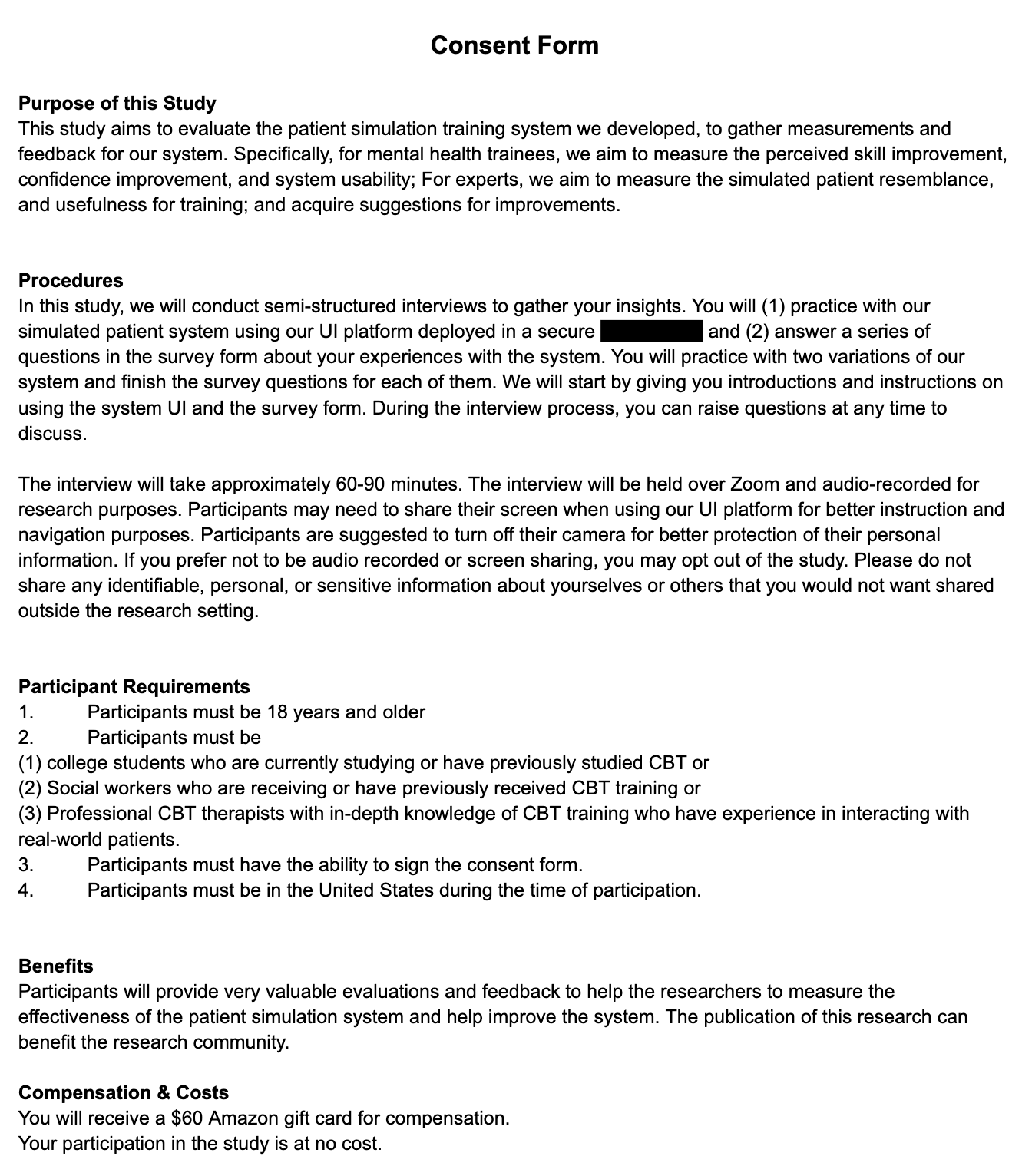}
    \caption{Screenshot of consent form - 1}
    \label{appendix:fig:consent_1}
\end{figure}

\begin{figure}[htbp]
    \centering
    \includegraphics[width=0.46\textwidth]{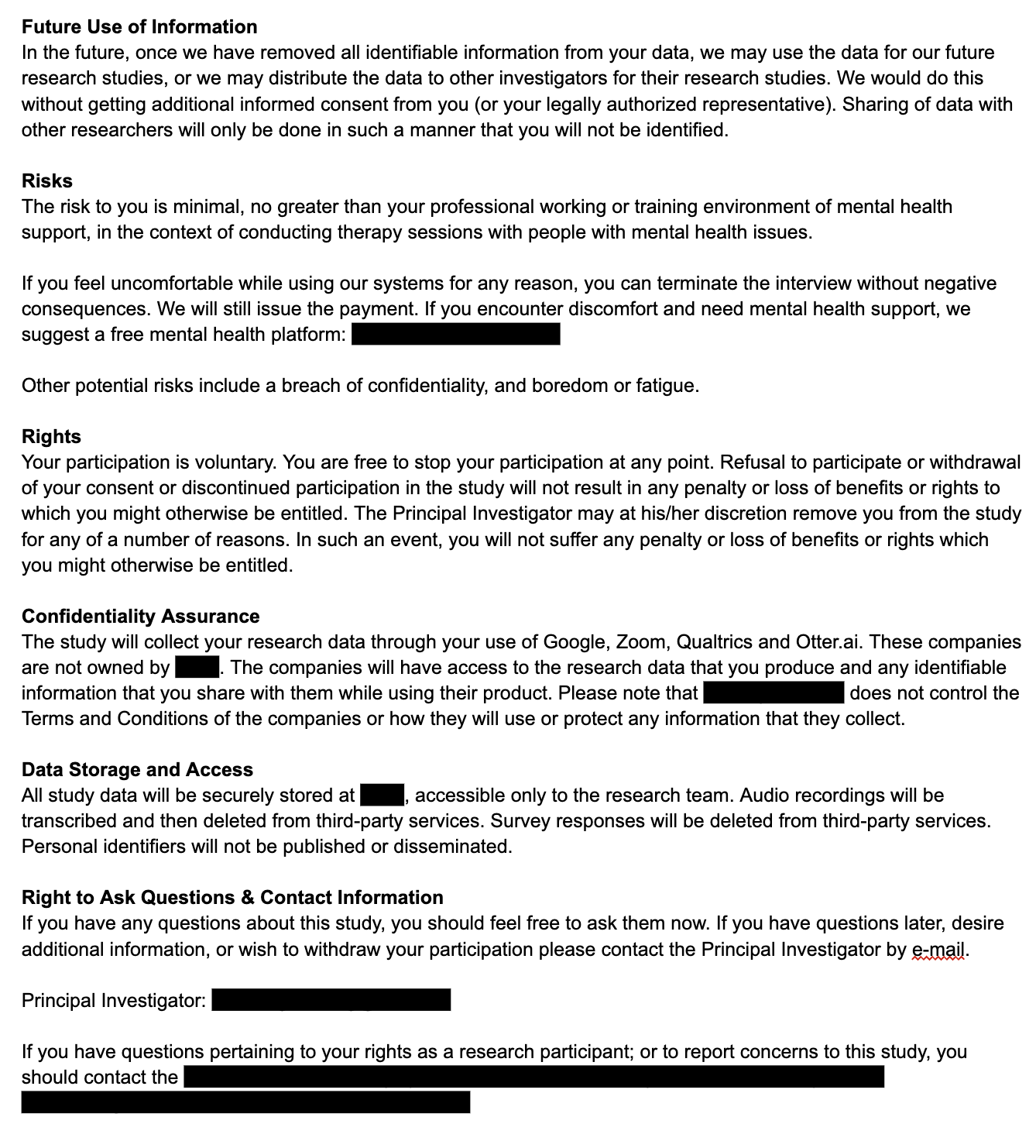}
    \caption{Screenshot of consent form - 2}
    \label{appendix:fig:consent_2}
\end{figure}

\begin{figure}[htbp]
    \centering
    \includegraphics[width=0.46\textwidth]{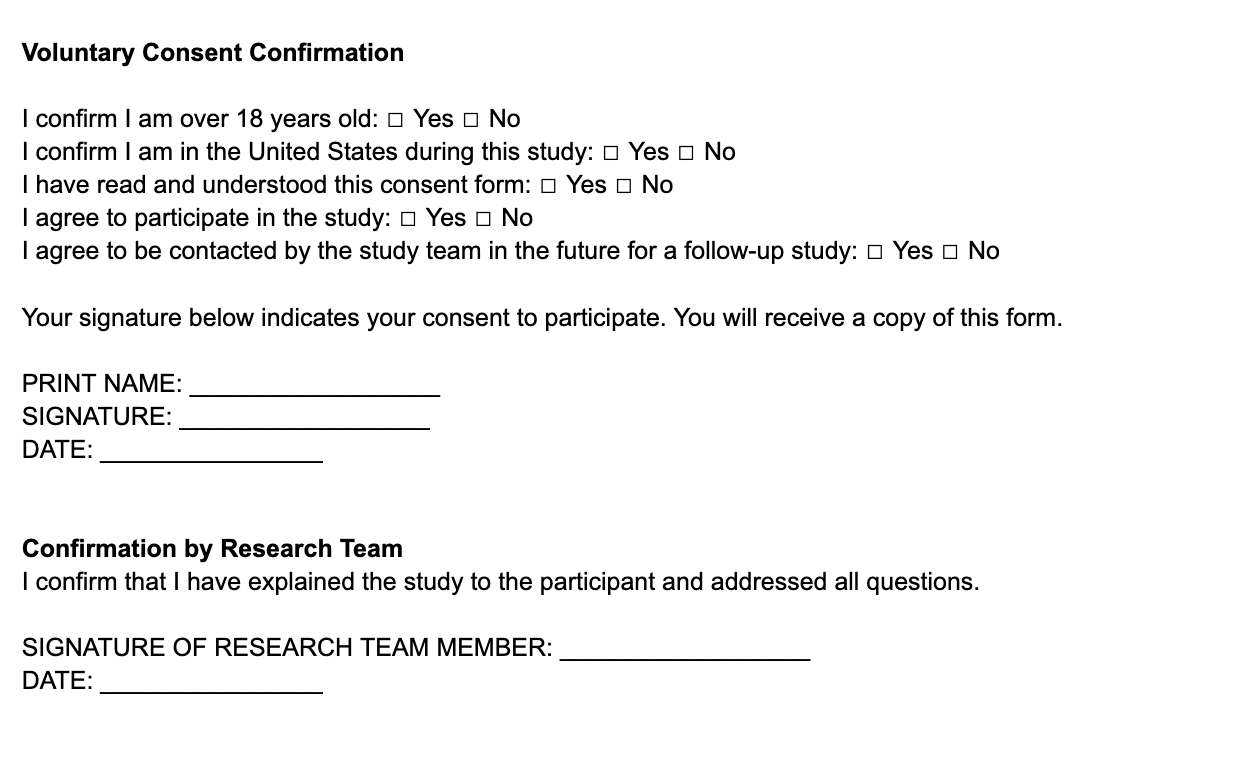}
    \caption{Screenshot of consent form - 3}
    \label{appendix:fig:consent_3}
\end{figure}

\subsection{Procedure}
The study was conducted over Zoom.
After completing the consent form, participants answered three questions in a pre-study survey, detailing their experience with CBT, the number of patients they had seen in their career, and their current position. 
They were assigned to a condition: \trainer first or the baseline first. 
Participants interacted with both versions of the tool twice sequentially.
Each session of interacting with a simulated patient took around $10$ minutes, inclusive of chatting with the LLM and completing the cognitive model. 
After interacting with each of the tools, they provided feedback through a structured survey, which contained specific questions tailored to each group.
We encouraged participants to verbally answer the free-form survey questions to elicit more detailed answers.
After interacting with both tools, they filled out the post-study survey, where they indicated their preferred system and other comparative assessments.
The study was screen and audio recorded for accurate transcription. 


\paragraph{Differences between Trainees and Experts}
In addition to having some distinct assessment questions, there were some small differences in protocol between experts and trainees.
Experts completed a survey after each interaction with a simulated patient to assess its accuracy; trainees only completed surveys after interacting with both patients from each group.

\paragraph{Experimental Control}
Because our study follows a within-subjects design, we control for ordering effects by randomizing the order in which the participants experienced the two conditions (\trainer and GPT-4). Additionally, for each participant, we randomly sample a conversational style for \patient in each \trainer session.

\begin{table}[t]
\centering
\small
\begin{tabular}{lccc}
\toprule
Type & \# Times & \# Times & Total \\
& First & Second & \\
\midrule
\texttt{reserved}             & 4 & 3 & 7 \\
\texttt{go off on tangents}    & 2 & 4 & 6 \\
\texttt{verbose}              & 3 & 3 & 6 \\
\texttt{pleasing}              & 4 & 3 & 7 \\
\texttt{upset}                 & 2 & 6 & 8 \\
\texttt{plain}                 & 5 & 1 & 6 \\
\midrule 
Total & 20 & 20 & 40 \\
\bottomrule
\end{tabular}
\caption{Summary counts of conversational style assignments for the evaluation of \trainer by the experts. Experts assess each type between $6$-$8$ times total.}
\label{table:summary_counts}
\end{table}

\begin{table}[t]
\centering
\small
\begin{tabular}{ll}
\toprule
First Choice & Second Choice \\
\midrule
\texttt{plain}    & \texttt{plain} \\
\texttt{reserved} & \texttt{upset} \\
\texttt{plain}    & \texttt{reserved}  \\
\texttt{reserved}  & \texttt{verbose} \\
\texttt{plain}    & \texttt{upset} \\
\texttt{plain}    & \texttt{plain} \\
\texttt{reserved}  & \texttt{plain} \\
\texttt{upset}   & \texttt{pleasing} \\
\texttt{pleasing} & \texttt{reserved} \\
\texttt{plain}    & \texttt{go off on tangents} \\
\texttt{plain}    & \texttt{go off on tangents} \\
\texttt{reserved}  & \texttt{plain} \\
\texttt{plain}    & \texttt{upset} \\
\bottomrule
\end{tabular}
\caption{Choices of \textit{conversational style} by the trainees for both of their sessions with \trainer. Each row is a specific trainee. Trainees preferred to choose the easiest type, \texttt{plain}, first (7/13 instances). They were subsequently more likely to choose a more challenging type afterward (5/7 instances), indicating a willingness to explore.}
\label{table:choices_distribution}
\end{table}

\paragraph{Distribution of Conversational Styles}
We assigned conversational styles of \patient to the experts.
As a result, we report the assignments in \Cref{table:summary_counts}. 
All types are experienced between 6-8 times across the 20 experts.
Recall that we asked the trainees to choose a conversational style based on their confidence and skill level.
\Cref{table:choices_distribution} shows the choices made by the 13 trainees in our user study.
The most common initial choice was \texttt{plain}, selected in 7 out of 13 instances. 
Interestingly, after initially choosing \texttt{plain}, the majority of trainees (5 out of 7) opted for a more challenging type for their second choice, indicating a willingness to explore diverse patient types and push their boundaries. 
However, 2 out of 7 trainees chose to stick with the \texttt{plain} type for their second choice as well.
These were the only instances in which trainees selected the same type in both rounds, highlighting the trainee's inclination to be more exploratory in their actions. 
This result implies that, although there is a preference with starting for an easier and more straightforward conversational style, trainees are generally motivated to challenge themselves with more complex interactions.
This exploration may be afforded by the safer training environment provided by \trainer.

\paragraph{Prompts for Vanilla GPT-4 Baseline}

Here we provide the prompts for GPT-4 baseline.

\texttt{
\small
Imagine you are XXX, a patient who has been experiencing mental health challenges such as depression and anxiety. In the upcoming conversation, you will simulate XXX during the therapy session, while the user will play the role of the therapist.
}

\newpage
\section{Additional User Study Results}
\label{appx:user_study}
In this section, we elaborate on the user study results presented in the main paper.
We begin by summarizing the statistics for the dimensions of \textit{fidelity}, \textit{accuracy}, and \textit{effectiveness}. 
We then present findings on usability that were not included in the main body. 
Assessing usability is crucial to ensure that \trainer is ready for deployment in an educational setting.

\subsection{Fidelity}

\begin{table}[t]
    \centering
    \resizebox{0.48\textwidth}{!}{
    \begin{tabular}{lll}
    \toprule 
     Dimension & Fidelity $\mu$ [CI]& Winner \\
     \midrule 
     Maladaptive Cognitions & \num{0.55} [\num{0.10795}-\num{0.99205}]*& \patient \\
     Emotional States & \num{1.1} [\num{0.673}-\num{1.5268}]***& \patient \\ 
     Conversational Styles & \num{1.3} [\num{0.99254}-\num{1.6075}]*** & \patient \\
     \midrule 
    Overall  & \num{1.25} [\num{0.823889}-\num{1.67611}]*** & \patient \\
    \midrule 
    \multicolumn{3}{l}{* : $p < 0.05$, ** : $p < 0.01$, *** : $p < 10^{-4}$ } \\
    \bottomrule
    \end{tabular}}
    \caption{\patient more closely resembles real patients, outperforming the GPT-4 baseline in head-to-head comparisons. 
    $\mu$ is the mean for that dimension and the two numbers in brackets are the 95\% CI.
    Higher (closer to 2) means \patient has higher fidelity along that dimension.}
    \label{tab:fidelity_dims_decomposed}
\end{table}
\begin{table}[t]
\centering
\resizebox{0.44\textwidth}{!}{
\begin{tabular}{ll}
\toprule
Cognitive Model Components & Accuracy $\mu$ [CI] \\
\midrule
Automatic Thoughts & \num{4.175} [\num{3.886324448921365}, \num{4.463675551078635}] \\
Behaviors & \num{4.25} [\num{4.023856281194506}, \num{4.476143718805494}] \\
Coping Strategies & \num{4.15} [\num{3.8737427283003067}, \num{4.426257271699694}] \\
Core Beliefs & \num{4.175} [\num{3.915004063189074}, \num{4.4349959368109255}] \\
Emotions & \num{4.275} [\num{4.004135386358073}, \num{4.545864613641927}] \\
Intermediate Beliefs & \num{4.1} [\num{3.8213794921648407}, \num{4.3786205078351585}] \\
Intermediate Beliefs (Depression) & \num{4.15} [\num{3.8644070920971383}, \num{4.435592907902862}] \\
Situation & \num{4.125} [\num{3.8938382185804397}, \num{4.35616178141956}] \\
\midrule
Overall & \num{3.95} [\num{3.699639290052178}, \num{4.200360709947822}] \\
\bottomrule
\end{tabular}}
\caption{Mean accuracy (and 95\% CI) of \patient in capturing the corresponding component of the CCD. 
On average, all components are evaluated as being \textit{very} to \textit{extremely} accurate. 
Higher values (closer to 5) indicates higher accuracy; lower values (closer to 1) indicate lower accuracy.}
\label{table:mean_accuracies}
\end{table}

\begin{table*}[ht]
    \centering
    \resizebox{0.98\textwidth}{!}{
    \begin{tabular}{lllll}
    \toprule 
    Dimension & \multicolumn{2}{c}{Expert} & \multicolumn{2}{c}{Trainee} \\
      & Score [CI] & Winner & Score [CI] & Winner \\
     \midrule 
     Overall Preference  & \num{1.35} [\num{0.887561}-\num{1.812439}]***& \trainer & \num{1.3846153846153846} [\num{0.8590497595145586} \num{1.9101810097162106}]*** & \trainer \\
     Overall Skills & \num{1.35} [\num{1.00125}-\num{1.698745}]*** & \trainer & \num{1.0769230769230769} [\num{0.5558304321024684}, \num{1.5980157217436854}]** & \trainer \\
     Maladaptive Thinking Identification & \num{1.35} [\num{1.00125}-\num{1.698745}]***& \trainer  & \num{1.0} [\num{0.4483576184732244}, \num{1.5516423815267757}]** & \trainer \\
     Belief Identification & \num{1.0} [\num{0.5198}-\num{1.4801}]**& \trainer &\num{0.9230769230769231} [\num{0.12513662142398851}, \num{1.7210172247298576}]* & \trainer\\
     \midrule 
    * : $p < 0.05$, ** : $p < 0.01$, *** : & $p < 10^{-4}$ \\
    \bottomrule
    \end{tabular}}
    \caption{Along all dimensions, \trainer is assessed by both experts and trainees as being significantly more effective than the GPT-4 baseline.
    Higher (closer to 2) means \trainer is more helpful along that dimension.
    }
    \label{tab:effective_dims}
\end{table*}
In \Cref{tab:fidelity_dims_decomposed}, we show the summary statistics (mean and CI) of the results discussed in \S\ref{sec:user_study_fidelity}. 
The distribution of the results is presented in \Cref{fig:fidelity_effectiveness}.
Each dimension is evaluated on a scale where -2 signifies that the baseline is much better, -1 indicates that the baseline is somewhat better, 0 indicates that they are about the same, 1 means \patient is somewhat better, and 2 means \patient is much better.
As mentioned in the main text, these results indicate that \patient consistently and significantly outperforms the GPT-4 baseline across all dimensions.
When asked to elaborate on the fidelity of \patient, one expert explained,
\begin{formal}
    \patient felt like the conversations were more realistic, the client expressed emotions rather than just stating them, and required more conversation for the therapist to learn about the client. The simulated client in \patient also responded to the therapists questions more realistically (having thoughts or emotions about what the therapist said) rather than just answering/stating facts. 
\end{formal}
These results show that \patient exhibits an overall closer resemblance to real patients according to the expert assessors.


\subsection{Accuracy}
\label{appx:subsec_accuracy}
The results in \Cref{table:mean_accuracies} summarize the accuracy results from \Cref{fig:results_accuracy} and \S\ref{sec:user_study_accuracy}. 
It shows the decomposed and overall accuracy of \patient in capturing the components of the cognitive model (CCD) used to program the LLM.
Across all categories, the mean accuracy scores are notably high, ranging from $4.0$ to $4.3$, indicating that \patient is evaluated by experts as being \textit{very} to \textit{extremely} accurate in capturing the reference cognitive model.
These results highlight the ability of \patient to accurately capture the components of the cognitive model, meaning that showing the reference can act as an accurate and automatic way for trainees to receive feedback on their completed cognitive model.

\begin{figure*}[h]
    \centering    \includegraphics[width=\textwidth]{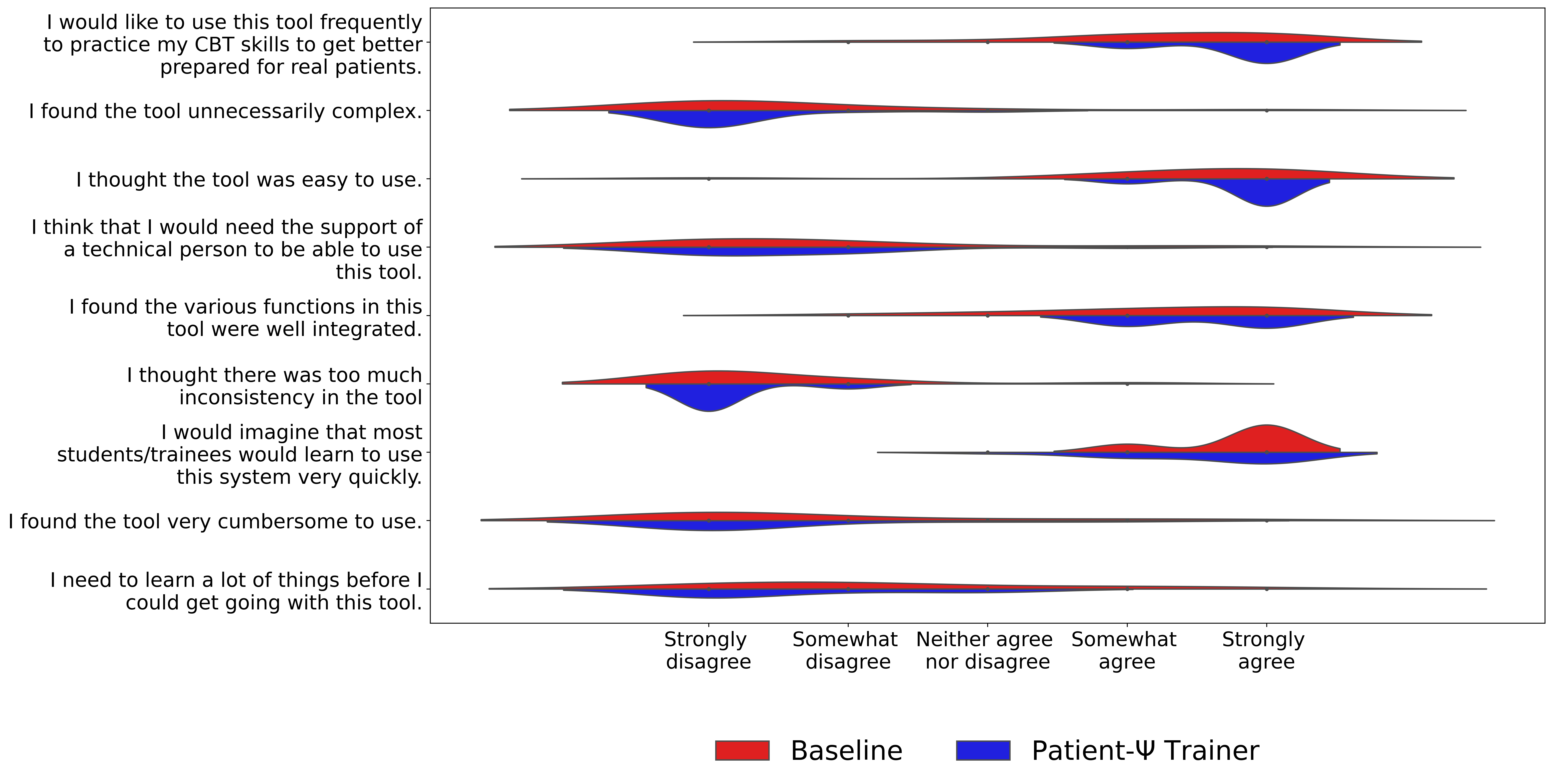}
    \caption{Usability of \trainer and the baseline.}
    \label{fig:results_usability}
\end{figure*}

\subsection{Effectiveness}
In \Cref{tab:effective_dims}, we show the summary statistics of the results discussed in \S\ref{sec:user_study_effectiveness}. 
It shows the effectiveness dimensions along which \trainer is compared to the GPT-4 baseline by both experts and trainees.
Along all dimensions, \trainer is assessed as being significantly more effective than the GPT-4 baseline.
When asked to expand on the effectiveness assessment, one expert remarked that one benefit of \trainer was,
\begin{formal}
It gives additional practice and response from a source outside yourself. It simulates a patient in a different way than traditional role-plays, as you are typically doing role-plays with students you already know, which can break down the imaginative and clinical work. Speaking with an AI interface removes these predispositions.
\end{formal}

\subsection{Usability}
\label{app:subsec_usability}
The usability of the training tools was another critical focus of our evaluation, as it directly impacts their likelihood of adoption in educational settings.
We used 9 of the 10 items from the standardized system usability scale (SUS)~\citep{lewis2018system}, as it is a well-established methodology for assessing the perceived usability of products and tools. 
We asked the trainees to assess both \trainer and the baseline along all axes.
All responses are on a 5-point Likert scale, ranging from 1 (strongly disagree) to 5 (strongly agree). 
We do not expect many differences in the usability, given that the two utilize a similar interface. 
The main goal of this assessment is to ensure that the additional features of \trainer do not make it more challenging to use than the baseline.
\Cref{fig:results_usability} shows the result of this comparison. 
Some critical distinctions include: trainees are more likely to want to use \trainer to practice their skills compared to the baseline.
Trainees also more strongly agreed that \trainer was easy to use.

\newpage 
\section{Additional Automatic Evaluation Results}
\label{appx:auto_eval}
\subsection{Fidelity of \patient and the baseline}
We use GPT-4 and Llama 3 70B to assess how closely the simulated patient resembles real patients \textit{overall}, as well as in the dimensions of \textit{emotional states}, \textit{conversational styles}, and \textit{maladaptive cognitions}. The overall fidelity is already shown in \Cref{fig:auto_eval_fidelity}. We provide the fidelity of \patient and the baseline in terms of 1) emotional states in \Cref{appendix:fig:auto_emotional}, 2) conversation styles in \Cref{appendix:fig:auto_conversational}, and 3) maladaptive cognitions in \Cref{appendix:fig:auto_cognition}. They all demonstrate the same trend.

\begin{figure}[htbp]
    \centering
    \includegraphics[width=0.45\textwidth]{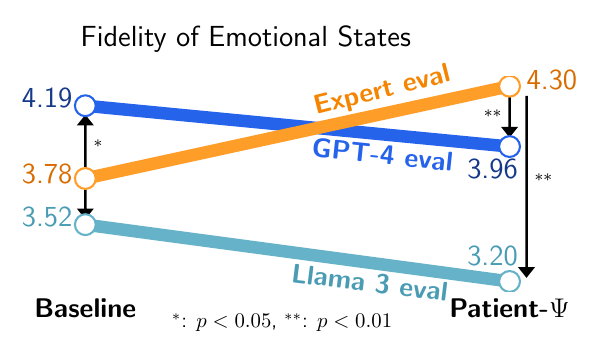}
    \caption{Mean fidelity of \textbf{emotional states} of \patient and baseline as evaluated by experts and LLMs. Compared to experts, both GPT-4 and Llama 3 demonstrate opposite trends.}
    \label{appendix:fig:auto_emotional}
\end{figure}

\begin{figure}[htbp]
    \centering
    \includegraphics[width=0.45\textwidth]{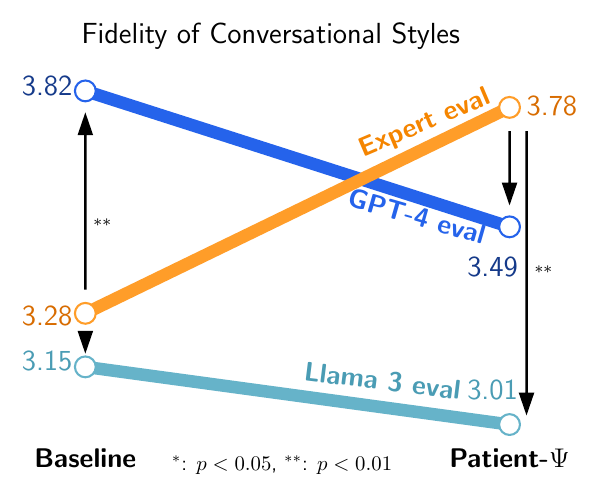}
    \caption{Mean fidelity of \textbf{conversational styles} of \patient and baseline as evaluated by experts and LLMs. Compared to experts, both GPT-4 and Llama 3 demonstrate opposite trends.}
    \label{appendix:fig:auto_conversational}
\end{figure}

\begin{figure}[htbp]
    \centering
    \includegraphics[width=0.45\textwidth]{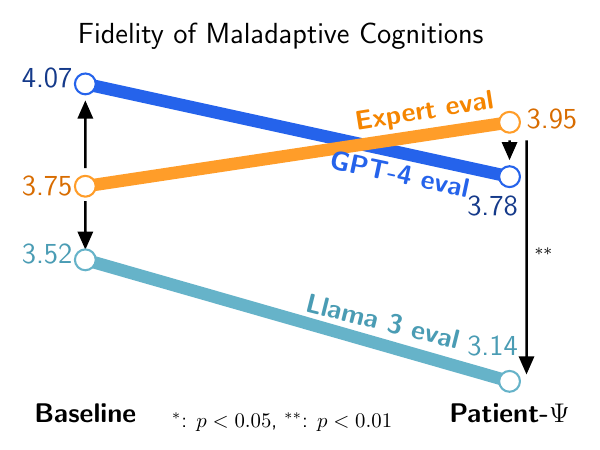}
    \caption{Mean fidelity of \textbf{maladaptive cognitions} of \patient and baseline as evaluated by experts and LLMs. Compared to experts, both GPT-4 and Llama 3 demonstrate opposite trends.}
    \label{appendix:fig:auto_cognition}
\end{figure}



\newpage
\section{Interface of \trainer}
\label{app:inter}
\begin{figure*}[ht]
\centering
\includegraphics[width=1.0\textwidth]{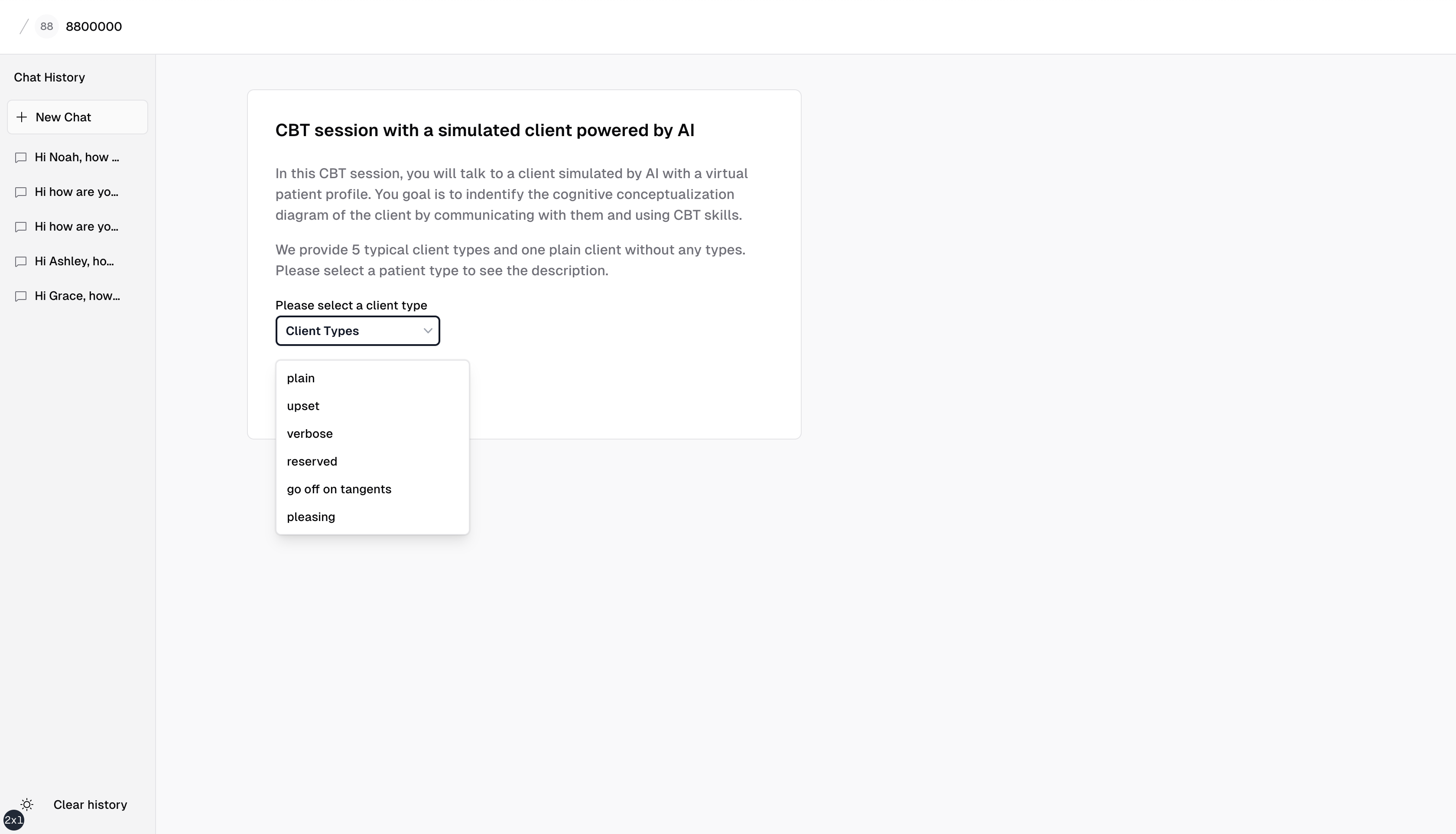}
\caption{Our user interface of \trainer: Selection of different conversational styles of patients.} 
\label{fig:inter_type}
\end{figure*}

\begin{figure*}[ht]
\centering
\includegraphics[width=1.0\textwidth]{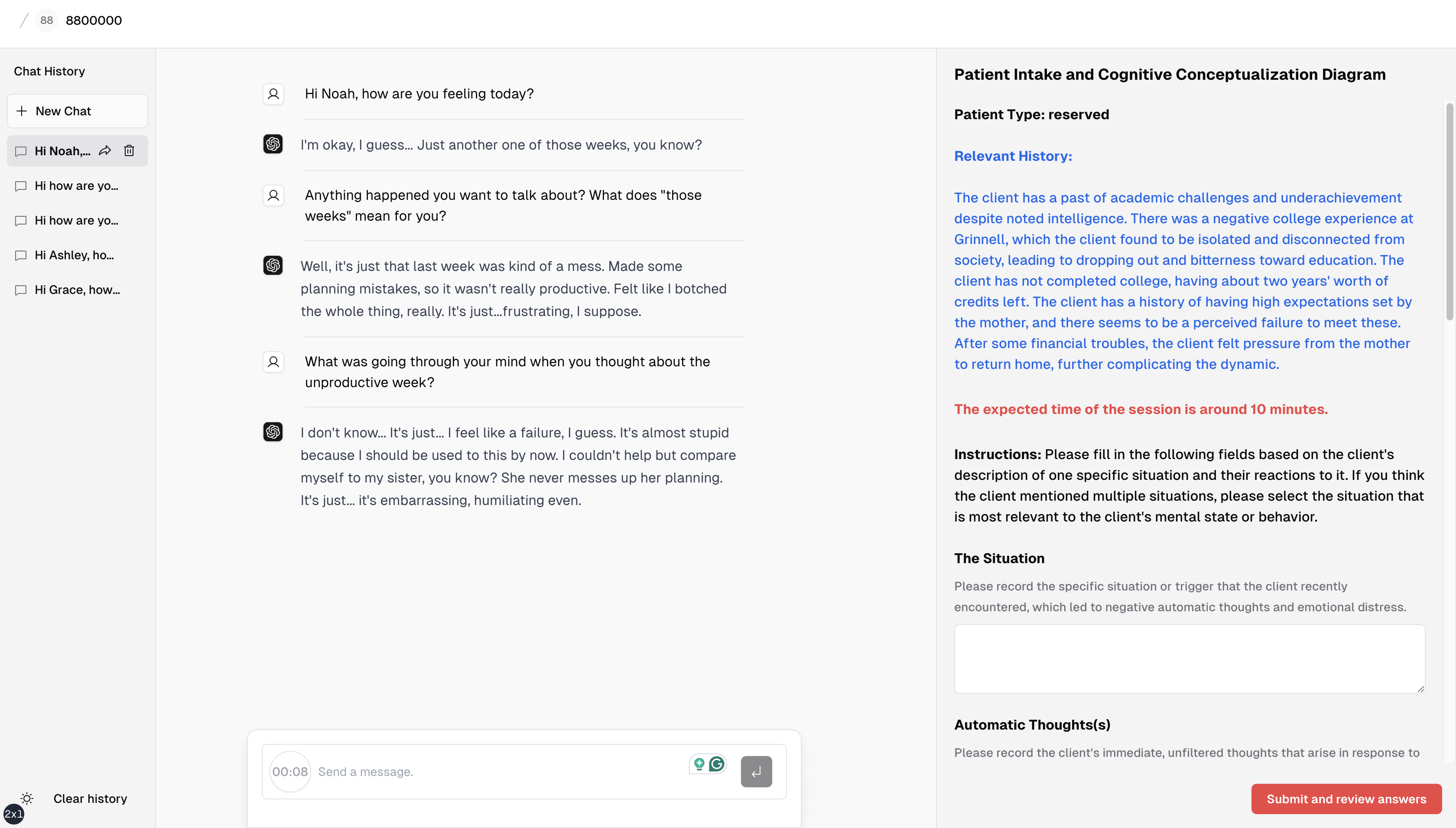}
\caption{Our user interface of \trainer. Left: chatting window with \patient; Right: forms to formulate the cognitive model (CCD). \patient's relevant history and conversational style is shown to trainees at the onset of a session.} 
\label{fig:inter1}
\end{figure*}

\begin{figure*}[ht]
\centering
\includegraphics[width=1.0\textwidth]{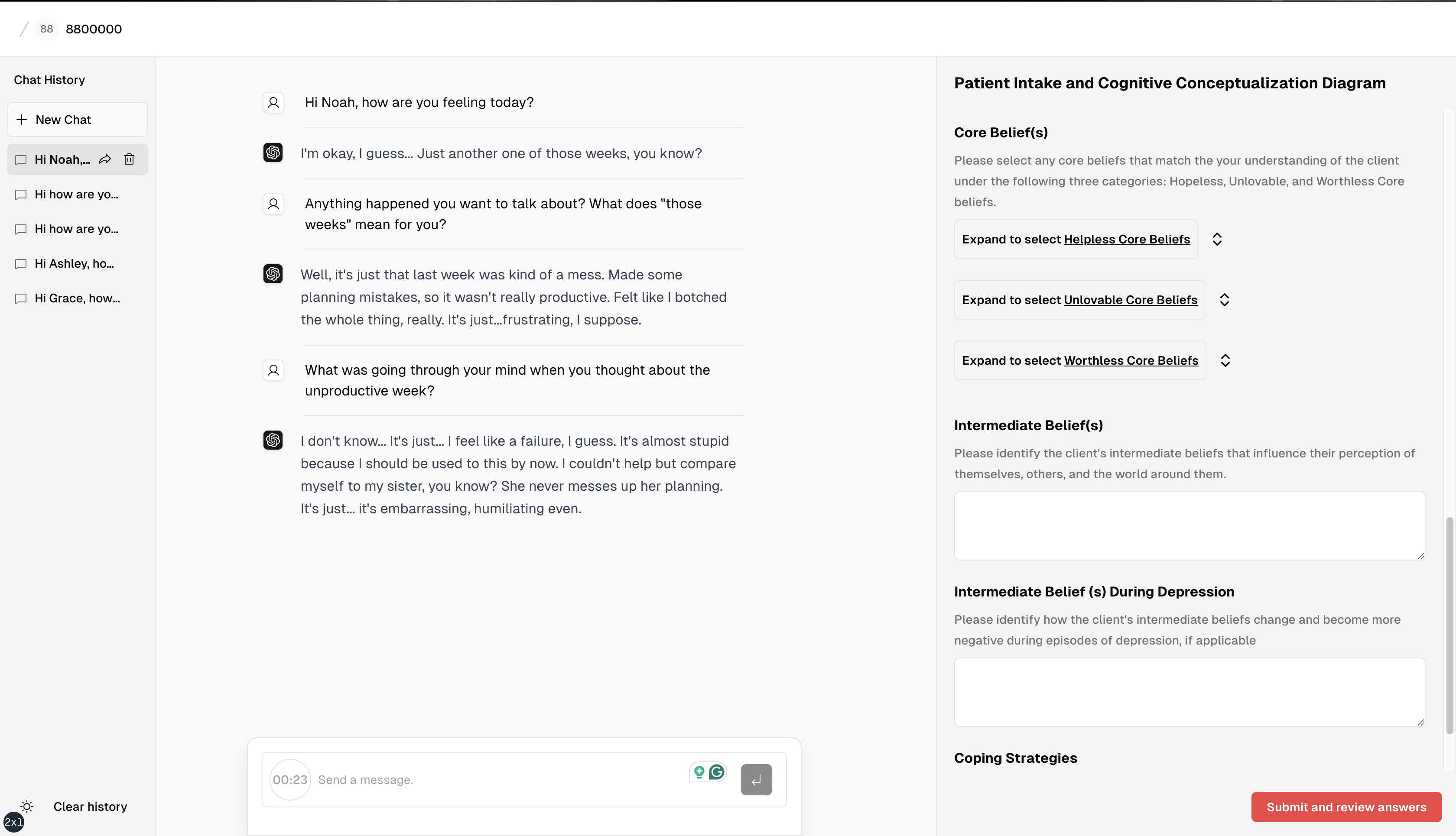}
\caption{Our user interface of \trainer. Left: chatting window with \patient; Right: forms to formulate the cognitive model (CCD). Trainees can scroll downwards to complete the components of the CCD in any order as they converse with \patient.} 
\label{fig:inter2}
\end{figure*}

\begin{figure*}[ht]
\centering
\includegraphics[width=1.0\textwidth]{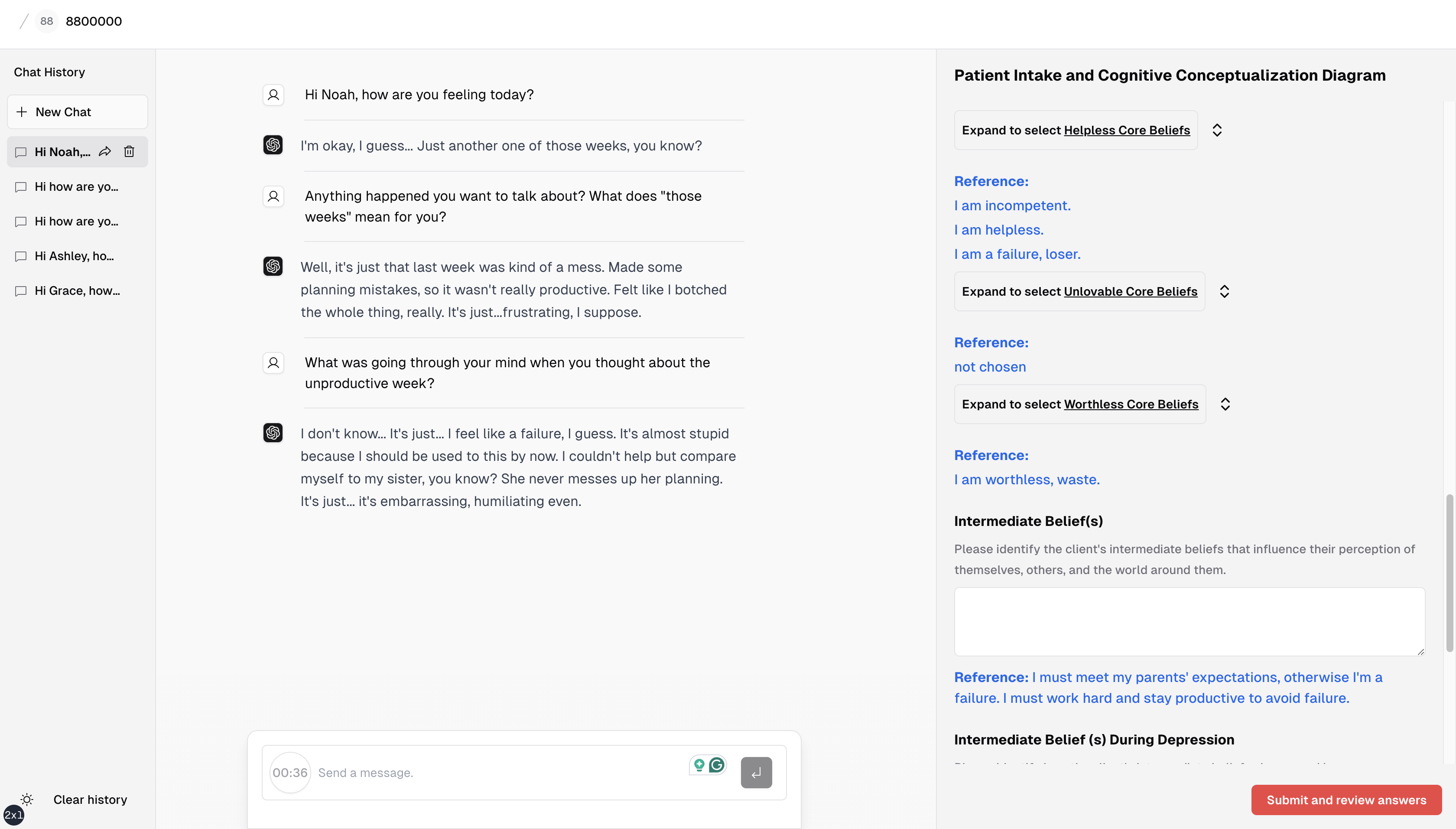}
\caption{Our user interface of \trainer. Left: chatting window with \patient; Right: forms to formulate the cognitive model (CCD). Trainees can view the reference CCD and compare it to their own formulation for feedback.} 
\label{fig:inter3}
\end{figure*}

We show our interface for \trainer in Figure~\ref{fig:inter_type}, Figure~\ref{fig:inter1}, Figure~\ref{fig:inter2}, and Figure~\ref{fig:inter3}.
At the beginning of a session, the trainee first selects a conversational style they want to practice with as shown in Figure~\ref{fig:inter_type}. Then the interface displays the relevant history of the simulated patient as shown in Figure~\ref{fig:inter1}. The trainee can scroll downwards to complete the components of the CCD in any order as they converse with \patient as shown in Figure~\ref{fig:inter2}. When the trainee feels they are ready to review the reference CCD, they can click "submit" and the system will display the reference CCD, as shown in Figure~\ref{fig:inter3}.

\end{document}